\documentclass[lettersize,journal]{IEEEtran}
\usepackage{amsmath,amssymb,amsfonts}
\usepackage{algorithmic}
\usepackage{algorithm}
\usepackage{array}
\usepackage[caption=false,font=normalsize,labelfont=sf,textfont=sf]{subfig}
\usepackage{textcomp}
\usepackage{stfloats}
\usepackage{url}
\usepackage{verbatim}
\usepackage{graphicx}
\usepackage{cite}
\usepackage{pifont}
\usepackage{threeparttable}
\usepackage{cite}

\usepackage[export]{adjustbox}

\usepackage{color}

\usepackage[table,xcdraw]{xcolor}

\usepackage{booktabs}

\usepackage{multirow}

\usepackage[table,xcdraw]{xcolor}

\usepackage[hidelinks]{hyperref}

\definecolor{linkcolor}{HTML}{ec008c}

\hypersetup{
	colorlinks=true,
	linkcolor=linkcolor,
	citecolor=linkcolor,
}

\usepackage{colortbl} 
\usepackage{xcolor} 
\usepackage{pgfplots}

\newcommand{\acrot}[1]{
 \pgfmathsetmacro\percentcolor{max(0,min(100,#1/1.5*100))} 
  \edef\tempcolor{\noexpand\cellcolor{red!\percentcolor}}
  \tempcolor #1 
}

\newcommand{\actrans}[1]{
 \pgfmathsetmacro\percentcolor{max(0,min(100,#1/3*100))}
  \edef\tempcolor{\noexpand\cellcolor{purple!\percentcolor}}
  \tempcolor #1 
}

\newcommand\liehat[1]{\left[ #1 \right]_\times}

\newcommand\mlcomment[1]{\iffalse #1 \fi}

\begin{document}

\title{RIs-Calib: An Open-Source Spatiotemporal Calibrator for Multiple 3D Radars and IMUs Based on Continuous-Time Estimation}

\author{
Shuolong Chen, Xingxing Li, Shengyu Li, Yuxuan Zhou, and Shiwen Wang

\thanks{
This work was supported by the National Key Research and Development Program of China under Grant 2023YFB3907100, the National Natural Science Foundation of China under Grant 423B240.
The authors are with the School of Geodesy and Geomatics (SGG), Wuhan University, Wuhan 430070, China.
Corresponding authors: Xingxing Li (\emph{xxli@sgg.whu.edu.cn}) and Shengyu Li (\emph{lishengyu@whu.edu.cn}).
}
\thanks{
The specific contributions of all authors to this work are listed in \emph{CRediT Authorship Contribution Statement} at the end of the article.
}
}

\markboth{Journal of \LaTeX\ Class Files,~Vol.~14, No.~8, August~2021}%
{Shell \MakeLowercase{\textit{et al.}}: A Sample Article Using IEEEtran.cls for IEEE Journals}


\maketitle

\begin{abstract}
		Aided inertial navigation system (INS), typically consisting of an inertial measurement unit (IMU) and an exteroceptive sensor, has been
		widely accepted as a feasible solution for navigation.
		Compared with vision-aided and LiDAR-aided INS, radar-aided INS could achieve better performance in adverse weather conditions since the radar utilizes low-frequency measuring signals with less attenuation effect in atmospheric gases and rain.
		For such a radar-aided INS, accurate spatiotemporal transformation is a fundamental prerequisite to achieving optimal information fusion.
		In this work, we present \emph{RIs-Calib}: a spatiotemporal calibrator for multiple 3D radars and IMUs based on continuous-time estimation, which enables accurate spatiotemporal calibration and does not require any additional artificial infrastructure or prior knowledge.
		Our approach starts with a rigorous initialization procedure to recover spatiotemporal parameters and kinematic B-splines of the sensor suite from the raw measurements.
		Following that, several batch optimizations would be conducted, where all parameters would be refined to global optimal states steadily.
		We validate and evaluate \emph{RIs-Calib} on both simulated and real-world experiments, and the results demonstrate that \emph{RIs-Calib} is capable of accurate and consistent calibration.
		We open-source our implementations at (\url{https://github.com/Unsigned-Long/RIs-Calib}) to benefit the research community.
\end{abstract}


\section{Introduction and Related Works}
	\IEEEPARstart{I}{nertial} navigation systems (INSs), utilizing inertial measurement units (IMUs) as sensing modality, can provide six-degrees-of-freedom motion estimation in the three-dimensional space.
	However, long-term drift typically exists in INSs due to the noise and biases in inertial measurements.
	A feasible solution to combat this issue is integrating exteroceptive sensors, such as a camera or light detection and ranging (LiDAR), into INSs, i.e., constructing aided INSs.
	While vision-aided INSs \cite{qin2018vins,song2023dgm} or LiDAR-aided INSs \cite{xu2021fast,dong2023pve} could achieve accurate ego-motion estimation, they are highly vulnerable to adverse weather, such as fog, rain, and snow.
	Conversely, radar-aided INSs \cite{doer2020radar,kramer2020radar,ng2021continuous} are insensitive to
	such challenging conditions as radars utilize lower-frequency signals which have lighter attenuation effect in atmospheric gases and rain \cite{zheng2020target}.
	Due to this fact, radar-aided INSs have attracted significant research efforts in recent years.
	For such systems, accurate spatiotemporal calibration is highly required since ill-calibrated spatiotemporal parameters could significantly affect the fusion performance \cite{chen2023targetless}.

	\begin{figure}[t]
		\centering
		\includegraphics[width=\linewidth]{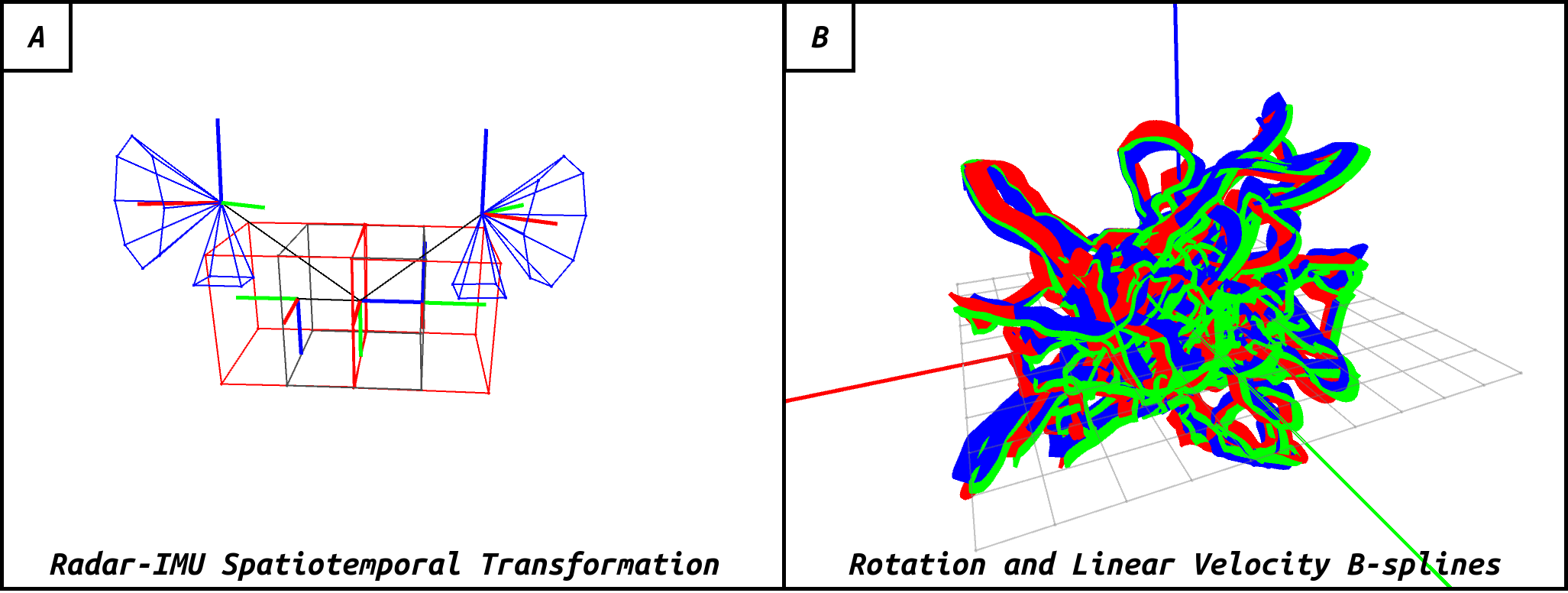}
		\caption{Runtime visualization of spatiotemporal calibration in real-world experiments in \emph{RIs-Calib}, where red boxes and bule entities in $A$ are IMUs and radars respectively, and discrete three-axis coordinate frames in $B$ present continuous-time B-splines maintained in estimator for spatiotemporal optimization.}
		\label{fig:mi_calib}
	\end{figure}
		
	For radar-related calibration, considering the structurelessness and sparsity of radar measurements, early works commonly employ specially designed artificial infrastructures as calibration targets to efficiently perform data association between heterogeneous sensors.
	Natour et al. \cite{el2015radar} utilized multiple painted metallic targets to determine extrinsics between a 2D radar and a camera, where the inter-target distance is also needed to provide prior constraints.
	By employing a radar detectable AR-marker, Song et al. \cite{song2017novel} adopted the paired point registration to calibrate spatial parameters for a 2D-Radar/Camera sensor suite.
	In addition to the camera, LiDAR is also a popular exteroceptive sensor to integrate with radar in autonomous vehicles to enhance perception performance.
	To determine extrinsics for a 3D LiDAR and a 2D radar, Per{\v{s}}i{\'c} et al. \cite{pervsic2017extrinsic} proposed a two-step calibration method, where a trihedral target with three orthogonal metal triangles is employed.
	Subsequently in \cite{pervsic2019extrinsic}, they considered an additional camera in the framework to perform joint calibration.
	Similarly, Domhof et al. \cite{domhof2021joint} designed an elaborate target with circular holes and a trihedral corner reflector to perform extrinsic calibration for a 2D radar, a camera, and a LiDAR.
	
	While 2D radars have been widely incorporated in autonomous vehicles (AVs), they only measure 2D information of targets, i.e., the distance, azimuth, and velocity in the planar coordinate.
	More advanced, 3D radars could provide additional elevation and thus have a wider sensing range, which have been progressively applied in AVs and other applications in recent years.
	Sch{\"o}ller et al. \cite{scholler2019targetless} employed a coarse-to-fine strategy and designed corresponding convolutional neural networks to perform the rotational calibration for a 3D radar and a camera.
	Different from its above target-based methods, this method is a targetless one, which requires no dedicated artificial targets during calibration, and thus with stronger flexibility.
	Similarly, as a targetless one, an extrinsic calibration method for 3D-Radar/Camera suites is proposed by Wise et al. \cite{wise2021continuous}, which utilizes the static natural targets in the environment as calibration targets to construct radar velocity constraints.
	Focus on online calibration, Doer et al. \cite{doer2020radar} presented a 3D radar inertial odometry, where extrinsics could be online optimized alongside other states in the estimator.
	However, prior knowledge about initial guesses of extrinsics is required to boot the odometry.
	
	In contrast to radar-related calibration, sufficient motion excitation is required in IMU-related calibration when collecting data to guarantee parameter observability.
	Benefiting from such dynamic calibration, it allows temporal determination.
	To calibrate the extrinsic, intrinsic, and temporal parameters of a Camera/IMU suite, Hu et al. \cite{hu2020multistate} employed a multi-state constrained invariant Kalman filter to perform online calibration, which greatly reduces the computational complexity compared with the traditional Kalman filter.
	Similarly, Huai et al. \cite{huai2022observability} proposed a keyframe-based visual-inertial odometry with online self-calibration, where a detailed observability analysis for spatiotemporal parameters is also performed. The above filter-based methods are discrete-time-based ones, which represents trajectories using discrete poses in estimator, thus simplified kinematic assumptions are generally needed when processing asynchronous measurements, which would introduce ineluctable errors in calibration.
	
	Different from the discrete-time trajectory representation, continuous-time representation employs continuous-time functions to encode trajectories, where poses can be computed at any arbitrary time, making it well-suited for asynchronous or high-frequency sensor fusion.
	The most representative work is the well-known \emph{Kalibr} \cite{furgale2013unified}, which first employs B-splines as the continuous-time trajectory representation to perform spatiotemporal calibration for an IMU and a global shutter camera.
	Subsequently, building upon \emph{Kalibr}, the rolling shutter camera is further supported in \cite{huai2022continuous} by Huai et al.
	Similarly employing B-splines as the continuous-time trajectory representation, Lv et al. \cite{lv2020targetless} proposed a targetless LiDAR/IMU calibration method. In their further work \cite{lv2022observability}, observability-aware modules are leveraged to address degenerate motions.
	
	The continuous-time estimation had also been employed in our previous work \cite{chen2023targetless,li2024targetless} for targetless camera/IMU and LiDAR/IMU spatiotemporal calibration.
	Different from them, the work presented in this paper focuses on 3D millimeter-wave radar/IMU spatiotemporal calibration, and faces several challenges compared with camera/IMU and LiDAR/IMU calibration: the structurelessness, sparsity, and significant noise of radar target measurements.
	This leads to difficulties in data association and continuous-time trajectory recovery if we still adhere to the pipeline of our previous work \cite{chen2023targetless,li2024targetless}.
	Considering this, we cleverly and reasonably abandon the previous practice of employing continuous-time representation to model the ego-motion trajectory (rotation and position curves), and instead employ it to model rotation and ego-velocity kinematic curves.
	Such modeling can rigorously and conveniently fuse both inertial measurements from IMUs and Doppler velocity measurements from radars in the estimator for spatiotemporal calibration, while avoiding challenging data association for radar measurements \cite{shi2023autonomous,dong2023new} and potential inconsistency and even association errors.
	
	In summary, while 3D radar-inertial navigation systems \cite{doer2020radar,kramer2020radar,ng2021continuous} have attracted significant research interest and have been increasingly developed recently, gaps exist in accurate spatiotemporal calibration for such sensor suites.
	To this end, based on continuous-time optimization, we propose a targetless spatiotemporal calibrator for sensor suites that integrates multiple 3D radars and IMUs.
	Specifically, we perform a rigorous initialization procedure to obtain initial guesses of states in the estimator, which requires no prior knowledge.
	Subsequently, based on the initialized parameters and raw measurements from radars and IMUs, we form a nonlinear factor graph including radar factors and IMU factors, and optimize it for several batches until the final convergence.
	The main contributions in our work can be listed as follows:
	\begin{enumerate}
		\item We propose a spatiotemporal calibration method for
		multiple 3D radars and IMUs based on continuous-time estimation, which supports accurate spatial, temporal, and intrinsic calibration, and requires no specially designed artificial targets or prior knowledge.
		
		\item Different from traditional continuous-time-based calibration methods that employ B-splines to represent time-varying pose (rotation and position), we innovatively employ them to encode rotation and velocity curves, which is naturally compatible with the measurements of both radar and IMU, and effectively reduces optimization complexity.
		
		\item We carried out both simulated and real-world experiments to demonstrate the high accuracy and repeatability of the proposed method. We open-source our implementations to benefit the community.
	\end{enumerate}
	
	\section{Preliminaries}
	\subsection{Notation}
	For a sensor suite that integrates $\mathcal{N}_b$ IMUs and $\mathcal{N}_r$ radars, we consider $\{b^i\}$ and $\{r^j\}$ as frames of the $i$-th IMU and the $j$-th radar respectively. The measurements of $\{b^i\}$ and $\{r^j\}$ are denoted as $\mathcal{D}(b^i)$ and $\mathcal{D}(r^j)$, where $i\in\left[0, \cdots, \mathcal{N}_b\right) $ and $j\in\left[0, \cdots, \mathcal{N}_r\right) $.
	We employ the Euclidean parametrization to represent the 3D transformation from frame $\{a\}$ to coordinate frame $\{b\}$ as follows:
	\begin{equation}
	\small
	\begin{aligned}
	{^{b}_{a}\boldsymbol{T}}=\begin{pmatrix}
	{^{b}_{a}\boldsymbol{R}}&{^{b}\boldsymbol{p}_{a}}\\\boldsymbol{0}_{1\times 3}&1
	\end{pmatrix}
	\end{aligned}
	\quad
	\mathrm{s.t.}
	\quad
	{^{b}_{a}\boldsymbol{R}}\in\mathrm{SO(3)},
	\; {^{b}\boldsymbol{p}_{a}}\in\mathbb{R}^3
	\end{equation}
	where ${^{b}_{a}\boldsymbol{T}}\in\mathrm{SE(3)}$ is the transform matrix.
	${^{b}_{a}\dot{\boldsymbol{R}}}\Leftrightarrow{^{b}_{a}\boldsymbol{\omega}}$,
	${^{b}_{a}\ddot{\boldsymbol{R}}}\Leftrightarrow{^{b}_{a}\dot{\boldsymbol{\omega}}}$, ${^{b}\dot{\boldsymbol{p}}_{a}}\Leftrightarrow{^{b}\boldsymbol{v}_{a}}$, and ${^{b}\ddot{\boldsymbol{p}}_{a}}\Leftrightarrow{^{b}\dot{\boldsymbol{v}}_{a}}$ are the angular velocity, angular acceleration, linear velocity, and linear acceleration of frame $\{a\}$ with respect to and parameterized in frame $\{b\}$, all of which live in $\mathbb{R}^3$.
	Finally, we denote $\hat{(\cdot)}$ as the noisy measurement, $\boldsymbol{\Sigma}_{(\cdot)}$ as the corresponding covariance matrix of a residual, and $\rho_{(\cdot)}$ as the Cauchy loss function to reduce the influence of outliers in least-squares problems.
	
	\subsection{Sensor Model}
	Adhering to the IMU model in \cite{lv2022observability}, we define the angular velocity and linear acceleration measurements of the $i$-th IMU at time $\tau$, i.e., ${^{g^i}}\hat{\boldsymbol{\omega}}(\tau)$ and ${^{a^i}}\hat{\boldsymbol{a}}(\tau)$ respectively, as:
	\begin{equation}
	\small
	\label{equ:imu_model}
	\begin{aligned}
	{^{g^i}}\hat{\boldsymbol{\omega}}(\tau)&
	={^{b^i}}\boldsymbol{\omega}(\tau)+\boldsymbol{b}_\omega^i+\boldsymbol{\varepsilon}_\omega^i
	=\boldsymbol{f}_{\omega}\left( {{^{b^i}}\boldsymbol{\omega}}(\tau),\boldsymbol{x}_{(b^i,in)}\right) 
	\\
	{^{a^i}}\hat{\boldsymbol{a}}(\tau)&
	={^{b^i}}\boldsymbol{a}(\tau)+\boldsymbol{b}_a^i+\boldsymbol{\varepsilon}_a^i
	=\boldsymbol{f}_{a}\left( {{^{b^i}}\boldsymbol{a}}(\tau),\boldsymbol{x}_{(b^i,in)}\right) 
	\end{aligned}
	\end{equation}
	with
	\begin{equation}
	\boldsymbol{x}_{\left(b^i,in \right) }=\left\lbrace 
	\boldsymbol{b}_\omega^i
	,\boldsymbol{b}_a^i
	\right\rbrace 
	\end{equation}
	where $\{g^i\}$ and $\{a^i\}$ are the sensor frames of the gyroscope and accelerometer, respectively;
	${{^{b^i}}\boldsymbol{\omega}}(\tau)$ and ${{^{b^i}}\boldsymbol{a}}(\tau)$ are the ideal angular velocity and linear	acceleration in frame $\{b^i\}$; $\boldsymbol{b}_\omega^i$ and $\boldsymbol{b}_a^i$ are the gyroscope and accelerometer biases, which are modeled as random walks and can be considered constant when the calibration data is short;
	$\boldsymbol{\varepsilon}_{\omega}^i$ and $\boldsymbol{\varepsilon}_{a}^i$ denote the measurement noises.
	
	In terms of the radar model, considering a natural target $\{t\}$ is tracked by the $j$-th radar $\{r^j\}$, we can obtain one radar measurement composed of the range $d$, azimuth $\theta$, elevation $\phi$, and radial velocity $v$ of the target with respect to the radar. These quantities are bound to each other as follows:
	\begin{equation}
	\label{equ:radar_model}
	\small
	v=
	\frac{{^{r^j}\boldsymbol{p}_t^\top}\cdot{^{r^j}\dot{\boldsymbol{p}}_t}}{d}
	\quad\mathrm{s.t.}\;
	{^{r^j}\boldsymbol{p}_t}=d\cdot\begin{pmatrix}
	\cos\theta\cos\phi
	\\
	\sin\theta\cos\phi
	\\
	\sin\phi
	\end{pmatrix}
	\end{equation}
	
	\subsection{Continuous-time Representation}
	To accurately fuse asynchronous and high-frequency measurements from multiple radars and IMUs, we employ the uniform B-splines as the continuous-time representation to encode the velocity and rotation trajectories.
	Compared to other continuous-time functions, e.g., Gaussian process \cite{nikolic2016non}, hierarchical wavelets \cite{anderson2014hierarchical}, and Chebyshev interpolation \cite{agrawal2024group}, the uniform B-splines have closed-from analytic derivatives and local controllability, which could yield a sparse system and reduce computational complexity in optimization.
	
	Specifically, given a sequence of velocity control points $\boldsymbol{v}_i$, $\boldsymbol{v}_{i+1}$, $\cdots$, $\boldsymbol{v}_{i+d}$ that are temporally uniformly distributed, the velocity B-spline of degree $d$ could be expressed as:
	\begin{equation}
	\small
	\label{equ:r3_bspline}
	\boldsymbol{v}(\tau)=\boldsymbol{v}_i+\sum_{j=1}^{d}
	\left( \boldsymbol{u}^\top\cdot
	\tilde{\boldsymbol{N}}^{d+1}_{j}\cdot
	\left( \boldsymbol{v}_{i+j}-\boldsymbol{v}_{i+j+1}\right) \right) 
	\end{equation}
	where $\tau\in[\tau_i,\tau_{i+1})$ is the time to interpolate the velocity; $\boldsymbol{u}^\top=\begin{pmatrix}1&u&\cdots&u^d\end{pmatrix}$ with $u=(\tau-\tau_i)/(\tau_{i+1}-\tau_i)$; $\tilde{\boldsymbol{N}}^{d+1}_{j}$ denotes the $j$-th column of matrix $\tilde{\boldsymbol{N}}^{d+1}$, which only depends on $d$.
	To balance the computational complexity and accuracy, we employ the cubic B-spline ($d=3$) and its corresponding matrix $\tilde{\boldsymbol{N}}^{(3+1)}$ can be found in \cite{litwostep}.
	
	As for the rotation B-spline, its control points which live in Lie Group $\mathrm{SO(3)}$ should be mapped to Lie Algebra $\mathfrak{so}(3)$ for scalar multiplication. Concretely, the $d$-degree rotation B-spline can be expressed as:
	\begin{equation}
	\small
	\label{equ:so3_bplsine}
	\boldsymbol{R}(\tau)=\boldsymbol{R}_i\cdot \prod_{j=1}^{d}\mathrm{Exp}\left( \boldsymbol{u}^\top\cdot
	\tilde{\boldsymbol{N}}^{d+1}_{j}\cdot
	\mathrm{Log}\left( \boldsymbol{R}_{i+j+1}^\top\cdot \boldsymbol{R}_{i+j}\right) \right) 
	\end{equation}
	where $\boldsymbol{R}_i$, $\boldsymbol{R}_{i+1}$, $\cdots$, $\boldsymbol{R}_{i+d}$ are a set of rotation control points;
	$\mathrm{Exp}(\cdot)$ is the operation that	mapping elements in $\mathfrak{so}(3)$ to $\mathrm{SO}(3)$, and $\mathrm{Log}(\cdot)$ is its inverse operation.
	
	\section{Methodology}
	\label{sect:Methodology}
	
	\begin{figure}[t]
		\centering
		\includegraphics[width=\linewidth]{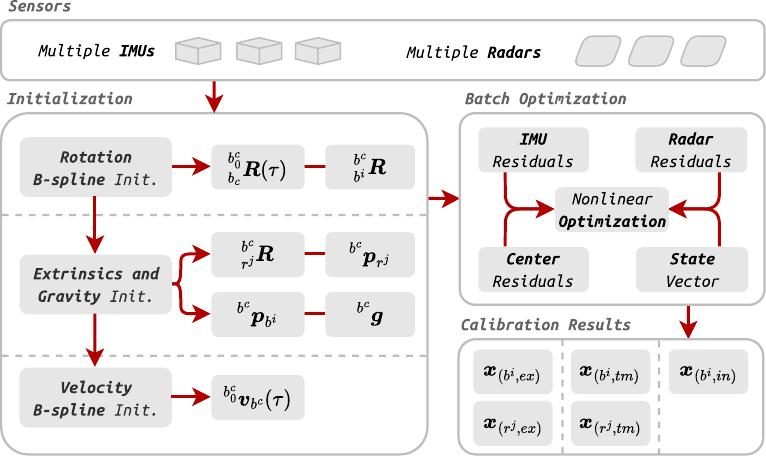}
		\caption{The pipeline of the proposed calibration method.}
		\label{fig:system}
	\end{figure}
	
	\subsection{Problem Formulation}
	The structure of the proposed spatiotemporal calibration method	for multiple radars and IMUs is shown in Fig. \ref{fig:system}.
	The system starts with a rigorous initialization procedure, which recovers extrinsics, world-frame gravity vector, as well as the velocity and rotation B-splines of a so-called virtual central IMU (namely the reference IMU, it's denoted as $\{b^c\}$).
	Subsequently, a nonlinear factor graph that minimizes IMU residuals and radar residuals would be optimized several batches until the final convergence.
	
	The full state vector $\boldsymbol{x}$ in the system includes extrinsics $\boldsymbol{x}_{\left(\;\cdot\;,\;ex \right) }$, time offsets $\boldsymbol{x}_{\left(\;\cdot\;,\;tm \right) }$, intrinsics $\boldsymbol{x}_{\left(\;\cdot\;,\;in\right) }$, a set of control points in the rotation and velocity B-splines of the central IMU $\boldsymbol{x}_{\left(cp,rot\right) }$ and $\boldsymbol{x}_{\left(cp,vel\right) }$, and the gravity vector $^{b^c_0}\boldsymbol{g}$ parameterized in the first frame of the central IMU $\{b^c_0\}$, which is defined as:
	\begin{equation}
	\small
	\begin{aligned}
	&\boldsymbol{x}=\left\lbrace 
	\begin{aligned}
	\cdots,
	&\boldsymbol{x}_{(b^i,ex ) },
	\boldsymbol{x}_{(b^i,tm ) },
	\boldsymbol{x}_{(b^i,in) },
	\cdots
	\\
	\cdots,
	&\boldsymbol{x}_{(r^j,ex ) },
	\boldsymbol{x}_{(r^j,tm ) },
	\cdots
	\\
	&\boldsymbol{x}_{(cp,rot) },
	\boldsymbol{x}_{(cp,vel) },
	{^{b^c_0}\boldsymbol{g}}
	\end{aligned}
	\right\rbrace
	\end{aligned}
	\end{equation}
	with
	\begin{equation}
	\small
	\begin{aligned}
	\boldsymbol{x}_{(b^i,ex )}&=\left\lbrace
	{^{b^c}_{b^i}\boldsymbol{R}},{^{b^c}\boldsymbol{p}_{b^i}}
	\right\rbrace
	&\boldsymbol{x}_{(b^i,tm )}&=\left\lbrace
	{^{b^c}\tau_{b^i}}
	\right\rbrace
	\\
	\boldsymbol{x}_{(r^j,ex )}&=\left\lbrace
	{^{b^c}_{r^j}\boldsymbol{R}},{^{b^c}\boldsymbol{p}_{r^j}}
	\right\rbrace 
	&\boldsymbol{x}_{(r^j,tm )}&=\left\lbrace
	{^{b^c}\tau_{r^j}}
	\right\rbrace
	\\
	\boldsymbol{x}_{(cp,rot) }&=\left\lbrace
	\cdots,{^{b^c_0}\boldsymbol{R}_k},\cdots 
	\right\rbrace
	&\boldsymbol{x}_{(cp,vel) }&=\left\lbrace
	\cdots,{^{b^c_0}\boldsymbol{v}_{k}},\cdots 
	\right\rbrace
	\end{aligned}
	\end{equation}
	where ${^{b^c}_{b^i}}\boldsymbol{R}$, ${^{b^c}}\boldsymbol{p}_{b^i}$, and ${^{b^c}\tau_{b^i}}$ are the extrinsic rotation, extrinsic translation, and time offset of the $i$-th IMU with respect to the central IMU, respectively;
	${^{b^c}_{r^j}}\boldsymbol{R}$, ${^{b^c}}\boldsymbol{p}_{r^j}$, and ${^{b^c}\tau_{r^j}}$ are the extrinsic rotation, extrinsic translation, and time offset of the $j$-th radar, respectively;
	The first frame of the reference IMU, i.e., coordinate frame $\{b^c_0\}$, is considered as the static world frame;
	${^{b^c_0}\boldsymbol{R}_k}$ and ${^{b^c_0}\boldsymbol{v}_{k}}$ are the $k$-th control points of the rotation and velocity B-splines respectively, with $k\in\left[0,\cdots,\mathcal{N}_{cp}\right)$;
	${^{b^c_0}\boldsymbol{g}}\in\mathbb{R}^3$ is the world-frame two-degrees-of-freedom gravity vector with a constant magnitude $\vert \vert{^{b^c_0}\boldsymbol{g}}\vert\vert\approx 9.81m/s^2$.
	Note that all spatiotemporal parameters are with respect to the virtual central IMU $\{b^c\}$, while other time-varying kinematic states are expressed with respect to $\{b^c_0\}$.
	
	\subsection{Initialization}
	The continuous-time-based spatiotemporal calibrator is a highly nonlinear system, which needs a rigorous initialization procedure to obtain a reasonable initial guess before performing the global optimization.
	Specifically, we initialize the B-splines of the virtual central IMU, extrinsics of sensors, and the gravity vector based on the raw measurements from multiple radars and IMUs.
	
	\subsubsection{Rotation B-spline Initialization}
	\label{sect:rot_bspline_init}
	We first perform a rotation-only B-spline fitting based on the raw angular velocity measurements of IMUs to recover the rotation B-spline, where the extrinsic rotations of IMUs as by-products could be initialized simultaneously.
	This is achieved by solving the following least-squares problem:
	\begin{equation}
	\small
	\label{equ:min_so3_bspline}
	\mathop{\min}\limits_{\boldsymbol{x}^\prime}
	\left\lbrace
	\sum_{\mathcal{N}_{b}}^{i}\sum_{\mathcal{D}(b^i)}^{k}
	\rho_\omega
	\left(\left\|
	\boldsymbol{r}^i_k(\omega)
	\right\|^2_{\boldsymbol{\Sigma}_\omega}\right)
	+\left\|
	\boldsymbol{r}_{ctr}(er)
	\right\|^2_{\boldsymbol{\Sigma}_{er}}
	\right\rbrace
	\end{equation}
	with
	\begin{equation}
	\small
	\label{equ:gyro_residual}
	\begin{aligned}
	&\boldsymbol{x}^\prime= 
	\left\lbrace \boldsymbol{x}_{(cp,rot)},
	\cdots,
	{^{b^c}_{b^i}{\boldsymbol{R}}}
	,\cdots
	\right\rbrace\;
	\boldsymbol{r}_{ctr}(er)=\sum^{i}_{\mathcal{N}_b}
	\mathrm{Log}\left( {^{b^c}_{b^i}\boldsymbol{R}}\right)
	\\
	&\boldsymbol{r}^i_k(\omega)=
	\boldsymbol{f}_{\omega}\left(
	{^{b^i}\boldsymbol{\omega}({^{b^c}\tau_k^i})},\boldsymbol{x}_{\left( b^i,in\right) }
	\right)-{^{g^i}}\hat{\boldsymbol{\omega}}(\tau_k^i)
	\end{aligned}
	\end{equation}
	where $\tau_k^i$ is the time of the $k$-th angular velocity measurement from the $i$-th IMU;
	${^{b^c}\tau_k^i}=\tau_k^i+{^{b^c}\tau_{b^i}}$ denotes the corresponding time stamped by the clock of the central IMU; $\boldsymbol{r}_{ctr}(er)$ is the so-called center residual for extrinsic rotation of IMUs to maintain a central rotation B-spline;  ${^{b^i}\boldsymbol{\omega}({^{b^c}\tau_k^i})}$ is the ideal angular velocity, which could be obtained by:
	\begin{equation}
	\small
	{^{b^i}\boldsymbol{\omega}(\tau)}=
	{^{b^c}_{b^i}\boldsymbol{R}^\top}\cdot
	{^{b^c_0}_{b^c}\boldsymbol{R}^\top(\tau)}\cdot
	{^{b^c_0}_{b^c}\dot{\boldsymbol{R}}(\tau)}
	\end{equation}
	where ${^{b^c_0}_{b^c}\boldsymbol{R}(\tau)}$ and ${^{b^c_0}_{b^c}\dot{\boldsymbol{R}}(\tau)}$ are the rotation and angular velocity of the central IMU at time $\tau$, which could be respectively computed by interpolating and differentiating the central rotation B-spline.
	
	\subsubsection{Extrinsics and Gravity Initialization}
	\label{sect:extri_grav_init}
	After initializing extrinsic rotations of IMUs and the rotation B-spline of the reference IMU, we move on to initialize extrinsic translations of IMUs, extrinsics of radars, and gravity. 
	Consider that the $j$-th radar takes one measurement of a static target  at time $\tau$, based on (\ref{equ:radar_model}) and subjected to the static constraint ${^{b^c_0}\dot{\boldsymbol{p}}_t}\equiv\boldsymbol{0}_{3\times 1}$, we have:
	\begin{equation}
	\label{equ:radar_velocity}
	\small
	v\cdot d=
	-{^{r^j}\boldsymbol{p}_t^\top}\cdot
	\underbrace{
		{^{b^c}_{r^j}\boldsymbol{R}^\top}\cdot
		{^{b^c_0}_{b^c}\boldsymbol{R}^\top(\tau)}\cdot
		{^{b^c_0}\dot{\boldsymbol{p}}_{r^j}(\tau)}
	}_{{^{b^c_0}\boldsymbol{v}^{r^j}_{r^j}(\tau)}}
	\end{equation}
	where ${^{b^c_0}\boldsymbol{v}^{r^j}_{r^j}(\tau)}$ denotes the velocity of frame $\{r^j\}$ with respect to frame $\{b^c_0\}$ but parameterized in frame $\{r^j\}$ at time $\tau$.
	By stacking multiple measurements in a single radar scan, ${^{b^c_0}\boldsymbol{v}^{r^j}_{r^j}(\tau)}$ could be solved analytically.
	
	Subsequently, we employ the velocity-level pre-integration to recover the gravity and uninitialized extrinsics based on the roughly solved radar velocities by (\ref{equ:radar_velocity}) and raw linear acceleration measurements from IMUs:	
	\begin{equation}
	\label{equ:gravity_init}
	\small
	\mathop{\min}\limits_{\boldsymbol{x}^{\prime\prime}}
	\left\lbrace 
	\sum_{\mathcal{N}_r}^{j}\sum_{\mathcal{D}(r^j)}^{k}\sum_{\mathcal{N}_b}^{i}
	\rho_v\left( 
	\left\|
	\boldsymbol{r}^{i,j}_{k,k+1}(v)  
	\right\|^2_{\boldsymbol{\Sigma}_v}\right) 
	+\left\|
	\boldsymbol{r}_{ctr}(et)
	\right\|^2_{\boldsymbol{\Sigma}_{et}}
	\right\rbrace 
	\end{equation}
	with
	\begin{equation}
	\small
	\begin{aligned}
	&\boldsymbol{x}^{\prime\prime}=\left\lbrace
	{^{b^c_0}\boldsymbol{g}},
	\cdots,
	\boldsymbol{x}_{(r^j,ex)},
	\cdots,
	{^{b^c}{\boldsymbol{p}_{b^i}}},
	\cdots
	\right\rbrace
	\quad
	\boldsymbol{r}_{ctr}(et)=\sum^{i}_{\mathcal{N}_b}
	{^{b^c}\boldsymbol{p}_{b^i}}
	\\
	&\boldsymbol{r}^{i,j}_{k,k+1}(v)=\mathcal{V}_{\tau_k,\tau_{k+1}}^{j}
	+\mathcal{V}_{\tau_k,\tau_{k+1}}^{i,\prime}\cdot{^{b^c}\boldsymbol{p}_{b^i}}
	-\mathcal{V}_{\tau_k,\tau_{k+1}}^{i,\prime\prime}
	\end{aligned}
	\end{equation}
	where $\boldsymbol{r}_{ctr}(et)$ is the center residual for extrinsic translation of IMUs; $\mathcal{V}_{\tau_k,\tau_{k+1}}^{j}$ denotes the velocity variation of the center IMU during $\left[\tau_k,\tau_{k+1} \right)$, which is derived from two consecutive scans of the $j$-th radar:
	\begin{equation}
	\label{equ:pre_integration_radar}
	\small
	\begin{aligned}
	\mathcal{V}_{\tau_k,\tau_{k+1}}^{j}&=
	{^{b^c_0}{\boldsymbol{v}}_{b^c}(\tau_{k+1})}-
	{^{b^c_0}{\boldsymbol{v}}_{b^c}(\tau_k)}
	-{^{b^c_0}\boldsymbol{g}}\cdot (\tau_{k+1}-\tau_k)
	\\
	{^{b^c_0}{\boldsymbol{v}}_{b^c}(\tau)}&={^{b^c_0}_{b^c}\boldsymbol{R}}(\tau)\cdot{^{b^c}_{r^j}\boldsymbol{R}}\cdot{^{b^c_0}{\boldsymbol{v}}^{r^j}_{r^j}(\tau)}
	+\liehat{{^{b^c_0}_{b^c}\boldsymbol{R}(\tau)}\cdot{^{b^c}{\boldsymbol{p}}_{r^j}}}
	{^{b^c_0}_{b^c}\dot{\boldsymbol{R}}(\tau)}
	\end{aligned}
	\end{equation}
	$\mathcal{V}_{\tau_k,\tau_{k+1}}^{i,\prime}$ and $\mathcal{V}_{\tau_k,\tau_{k+1}}^{i,\prime\prime}$ are quantities obtained by numerical integration based on raw linear acceleration measurements and recovered states in least-squares problem (\ref{equ:min_so3_bspline}):
	\begin{equation}
	\small
	\begin{aligned}
		\mathcal{V}_{\tau_k,\tau_{k+1}}^{i,\prime}&=
	\int_{\tau_k}^{\tau_{k+1}}
	{\left(\liehat{{^{b^c_0}_{b^c}\ddot{\boldsymbol{R}}(\tau)}}
		+\liehat{{^{b^c_0}_{b^c}\dot{\boldsymbol{R}}(\tau)}}^2\right) \cdot{^{b^c_0}_{b^c}\boldsymbol{R}(\tau)}}
	\cdot d\tau
	\\
	\mathcal{V}_{\tau_k,\tau_{k+1}}^{i,\prime\prime}&=
	\int_{\tau_k}^{\tau_{k+1}}
	{{^{b^c_0}_{b^c}\boldsymbol{R}(\tau)}\cdot{^{b^c}_{b^i}\boldsymbol{R}}\cdot{^{b^i}\boldsymbol{a}(\tau)}}
	\cdot d\tau
	\end{aligned}
	\end{equation}
	
	\subsubsection{Velocity B-spline Initialization}
	\label{sect:vel_bspline_init}
	With the initialized gravity vector, we recover the velocity B-spline and refine the quantities obtained in the previous step based on the raw measurements from radars and accelerometers.
	The corresponding least-squares problem could be expressed as:
	\begin{equation}
	\label{equ:vel_bspline}
	\small
	\mathop{\min}\limits_{\boldsymbol{x}^{\prime\prime\prime}}
	\left\lbrace 
	\begin{aligned}
	&\sum_{\mathcal{N}_r}^{j}\sum_{\mathcal{D}(r^j)}^{k}\sum_{\mathcal{N}_b}^{i}
	\rho_v\left( 
	\left\|
	\boldsymbol{r}^{i,j}_{k,k+1}(v)
	\right\|^2_{\boldsymbol{\Sigma}_v}\right) +
	\left\|
	\boldsymbol{r}_{ctr}(et)
	\right\|^2_{\boldsymbol{\Sigma}_{et}}
	\\
	&+
	\sum_{\mathcal{N}_r}^{j}\sum_{\mathcal{D}(r^j)}^{k}\sum_{\mathcal{D}(r^j,k)}^{l}
	\rho_r\left( 
	\left\|
	\boldsymbol{r}^{j}_{k,l}(r)
	\right\|^2_{\boldsymbol{\Sigma}_r}\right) 
	\end{aligned}
	\right\rbrace 
	\end{equation}
	with
	\begin{equation}
	\small
	\begin{aligned}
	\boldsymbol{x}^{\prime\prime\prime}&=\left\lbrace
	\boldsymbol{x}_{(cp,vel)},
	{^{b^c_0}\boldsymbol{g}},
	\cdots,
	\boldsymbol{x}_{(r^j,ex)},
	\cdots,
	{^{b^c}{\boldsymbol{p}_{b^i}}},
	\cdots
	\right\rbrace
	\\
	\boldsymbol{r}^{j}_{k,l}(r)&=
	v^{j}_{k,l}\cdot d^{j}_{k,l}+
	{^{r^j}\boldsymbol{p}_{t^{j}_{k,l}}^\top}
	{^{b^c}_{r^j}\boldsymbol{R}^\top}
	{^{b^c_0}_{b^c}\boldsymbol{R}^\top({^{b^c}\tau^{j}_{k,l}})}\cdot
	{^{b^c_0}\boldsymbol{v}_{r^j}({^{b^c}\tau^{j}_{k,l}})}
	\end{aligned}
	\end{equation}
	where $\mathcal{D}(r^j,k)$ denotes the $k$-th scan of the $j$-th radar; $\{v^{j}_{k,l},d^{j}_{k,l},{^{r^j}\boldsymbol{p}_{t^{j}_{k,l}}}\}$ is the $l$-th measurement in $\mathcal{D}(r^j,k)$, and $\tau^{j}_{k,l}$ is its timestamp; 
	${^{b^c}\tau^{j}_{k,l}}=\tau^{j}_{k,l}+{^{b^c}\tau_{r^j}}$ is the corresponding measuring time stamped by the clock of the central IMU;
	${^{b^c_0}\boldsymbol{v}_{r^j}({^{b^c}\tau^{j}_{k,l}})}$ is the velocity of the $j$-th radar, which could be obtained by:
	\begin{equation}
	\small
	{^{b^c_0}\boldsymbol{v}_{r^j}(\tau)}=
	{^{b^c_0}{\boldsymbol{v}}_{b^c}(\tau)}-
	\liehat{{^{b^c_0}_{b^c}\boldsymbol{R}(\tau)}\cdot
		{^{b^c}{\boldsymbol{p}}_{r^j}}}
	{^{b^c_0}_{b^c}\dot{\boldsymbol{R}}(\tau)}
	\end{equation}
	where ${^{b^c_0}{\boldsymbol{v}}_{b^c}(\tau)}$ is the velocity of the central IMU at time $\tau$, which can be interpolated from the velocity B-spline.
	Note that we use the velocity-level pre-integration residual again when organizing this least-squares problem.
	The difference from the previous step is that the velocities of the central IMU, i.e., ${^{b^c_0}{\boldsymbol{v}}_{b^c}(\tau_k)}$ and ${^{b^c_0}{\boldsymbol{v}}_{b^c}(\tau_{k+1})}$ in $\mathcal{V}_{\tau_k,\tau_{k+1}}^{j}$, are obtained by directly interpolating the velocity B-spline, rather than by the radar velocities from a pre-solved linear least-squares problem.
	
	For the left states in the estimators, such as intrinsics of IMUs, they are set as zeros or identities in initialization.
	At this point, the initialization procedure is completed.
		
	\subsection{Batch Optimization}
	\label{sect:batch_opt}
	After initialization, we form and solve a nonlinear least-squares problem by minimizing IMU residuals, radar residuals, and center residuals.
	
	\subsubsection{IMU Residual}
	The IMU residual is composed of gyroscope residual and accelerometer residual, in which the gyroscope residual for the $k$-th measurement of the $i$-th IMU has been defined as $\boldsymbol{r}^i_k(\omega)$ in (\ref{equ:min_so3_bspline}). As for the accelerometer residual, we define it as:
	\begin{equation}
	\small
	\boldsymbol{r}^i_k(a)=
	\boldsymbol{f}_{a}\left(
	{^{b^i}\boldsymbol{a}({^{b^c}\tau_k^i})},\boldsymbol{x}_{\left( b^i,in\right) }
	\right)-{^{a^i}}\hat{\boldsymbol{a}}({\tau_k^i})
	\end{equation}
	where ${^{b^i}\boldsymbol{a}({^{b^c}\tau_k^i})}$ is the ideal linear acceleration from the velocity B-spline:
	\begin{equation}
	\small
	{^{b^i}\boldsymbol{a}(\tau)}=
	{^{b^c}_{b^i}\boldsymbol{R}^\top}\cdot
	{^{b^c_0}_{b^c}\boldsymbol{R}^\top(\tau)}\cdot
	({^{b^c_0}\dot{\boldsymbol{v}}_{b^i}(\tau)}-{^{b^c_0}\boldsymbol{g}})
	\end{equation}
	with
	\begin{equation}
	\small
	\begin{aligned}
	{^{b^c_0}\dot{\boldsymbol{v}}_{b^i}(\tau)}&=
	{^{b^c_0}\dot{\boldsymbol{v}}_{b^c}(\tau)}-\liehat{{^{b^c_0}_{b^c}\boldsymbol{R}(\tau)}\cdot{^{b^c}\boldsymbol{p}_{b^i}}}
	{^{b^c_0}_{b^c}\ddot{\boldsymbol{R}}(\tau)}-
	\\&
	\hspace{4mm}\liehat{{^{b^c_0}_{b^c}\dot{\boldsymbol{R}}(\tau)}}
	\liehat{{^{b^c_0}_{b^c}\boldsymbol{R}(\tau)}\cdot{^{b^c}\boldsymbol{p}_{b^i}}}{^{b^c_0}_{b^c}\dot{\boldsymbol{R}}(\tau)}
	\end{aligned}
	\end{equation}
	
	\subsubsection{Radar Residual}
	The radar residual in batch optimization is the same as $\boldsymbol{r}^{j}_{k,l}(r)$ in (\ref{equ:vel_bspline}).
	
	\subsubsection{Center Residual}
	Since we introduce a virtual central IMU and maintain its B-splines in the estimator, the center residuals are required to ensure the system has a unique least-squares solution.
	Adhering to our previous work \cite{chen2023targetless}, we construct three types of center residuals, i.e.,
	the rotational center residual, the translational center residual,
	and the temporal center residual.
	The rotational and translational center residuals have been defined in (\ref{equ:min_so3_bspline}) and (\ref{equ:gravity_init}) as $\boldsymbol{r}_{ctr}(er)$ and $\boldsymbol{r}_{ctr}(et)$, respectively.
	In terms of the temporal center residual, we define it as:
	\begin{equation}
	\small
	\boldsymbol{r}_{ctr}(tm)=\sum^{i}_{\mathcal{N}_b}
	{^{b^c}\boldsymbol{\tau}_{b^i}}
	\end{equation}
	
	Finally, we stack all residuals and describe the batch optimization problem as the following nonlinear least-squares problem:
	\begin{equation}
	\small
	\mathop{\min}\limits_{\boldsymbol{x}}
	\left\lbrace
	\begin{aligned}
	&\sum_{\mathcal{N}_b}^{i}\sum_{\mathcal{D}(b^i)}^{k}
	\left( \rho_\omega
	\left(\left\|
	\boldsymbol{r}^i_k(\omega)
	\right\|^2_{\boldsymbol{\Sigma}_\omega}\right)+
	\rho_a\left(\left\|
	\boldsymbol{r}^i_k(a)
	\right\|^2_{\boldsymbol{\Sigma}_a}\right)\right)
	\\
	&+\sum_{\mathcal{N}_r}^{j}\sum_{\mathcal{D}(r^j)}^{k}\sum_{\mathcal{D}(r^j,k)}^{l}
	\rho_r\left( 
	\left\|
	\boldsymbol{r}^{j}_{k,l}(r)
	\right\|^2_{\boldsymbol{\Sigma}_r}\right) 
	\\
	&+\left\|\boldsymbol{r}_{ctr}(er)
	\right\|^2_{\boldsymbol{\Sigma}_{er}}
	+\left\|\boldsymbol{r}_{ctr}(et)
	\right\|^2_{\boldsymbol{\Sigma}_{et}}
	+\left\|\boldsymbol{r}_{ctr}(tm)
	\right\|^2_{\boldsymbol{\Sigma}_{tm}}
	\end{aligned}
	\right\rbrace
	\end{equation}
	We employ the \emph{Ceres solver} \cite{ceres_solver} to solve this problem.
		
		\section{Simulation}
		
		\begin{figure}[t]
			\centering
			\includegraphics[width=\linewidth]{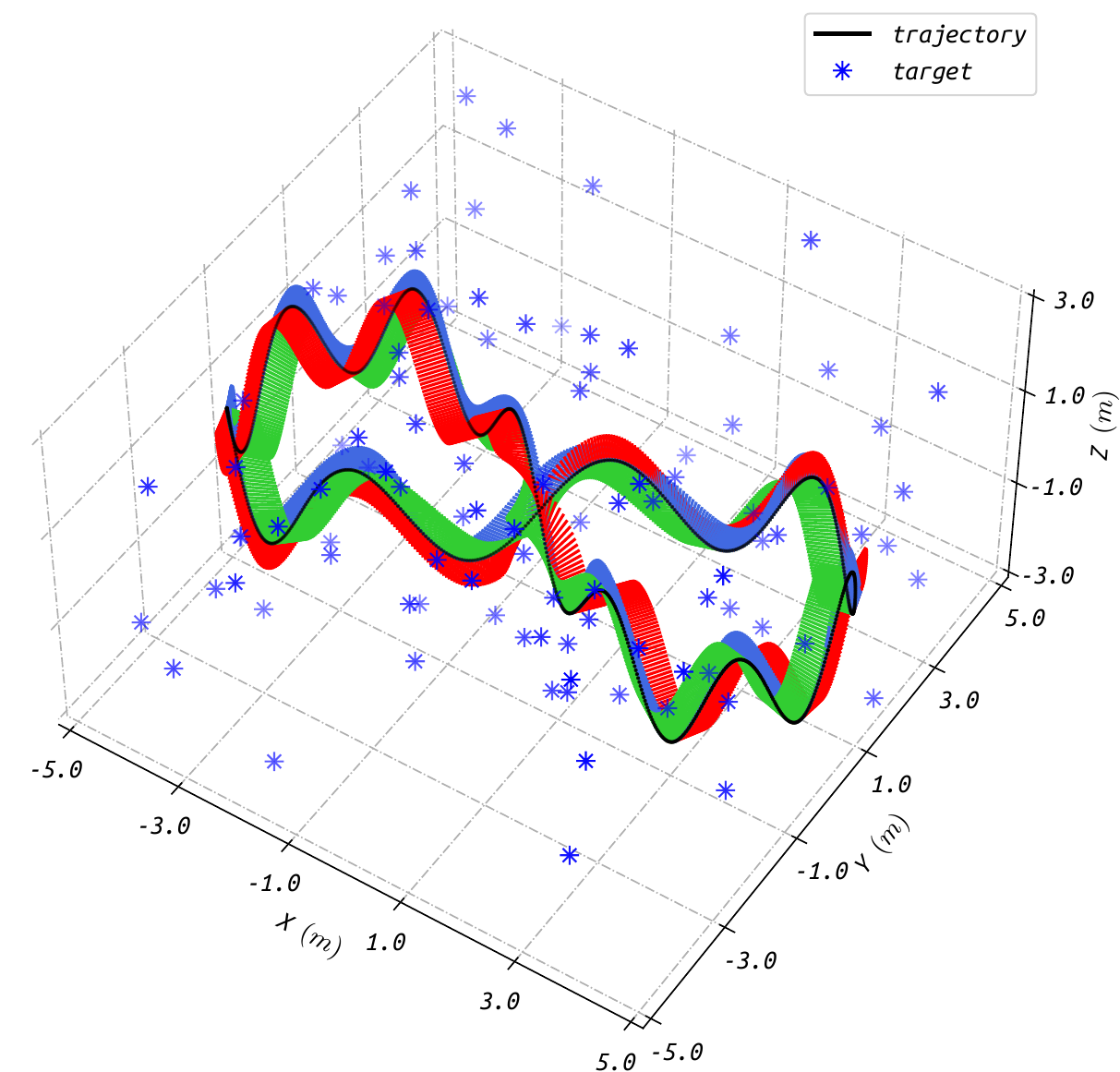}
			\caption{The simulated scenario with uniformly distributed static targets and a sufficiently excited 8-shape trajectory represented by discrete three-axis coordinates.}
			\label{fig:simu_scene}
		\end{figure}
		
		\begin{figure*}[t]
			\centering
			\includegraphics[width=\linewidth]{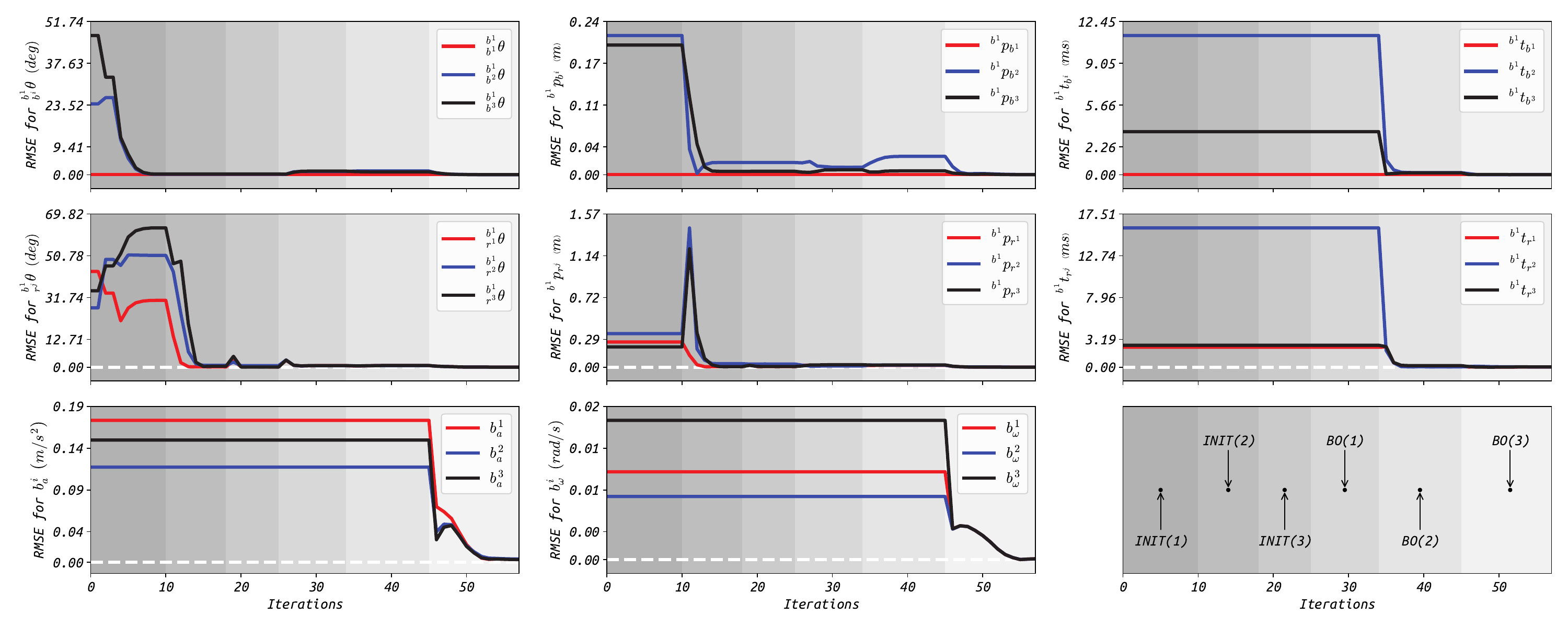}
			\caption{The system state and convergence performance in different stages: ($i$) \emph{INIT(1)}: Rotation B-spline Initialization (see Section \ref{sect:rot_bspline_init}); ($ii$) \emph{INIT(2)}: Extrinsics and Gravity Initialization (see Section \ref{sect:extri_grav_init}); ($iii$) \emph{INIT(2)}: Velocity B-spline Initialization (see Section \ref{sect:vel_bspline_init}); $(iv)$ \emph{BO(j)}: the $j$-th batch optimization (see Section \ref{sect:batch_opt}). The white dashed line in each sub-figure indicates the zero line. For better readers' understanding, all spatiotemporal parameters are with respect to \emph{IMU-1}, rather than with respect to the virtual central IMU maintained in the estimator.}
			\label{fig:simu_opt_process}
		\end{figure*}
		
		\begin{figure}[t!]
			\centering
			\includegraphics[width=0.9\linewidth]{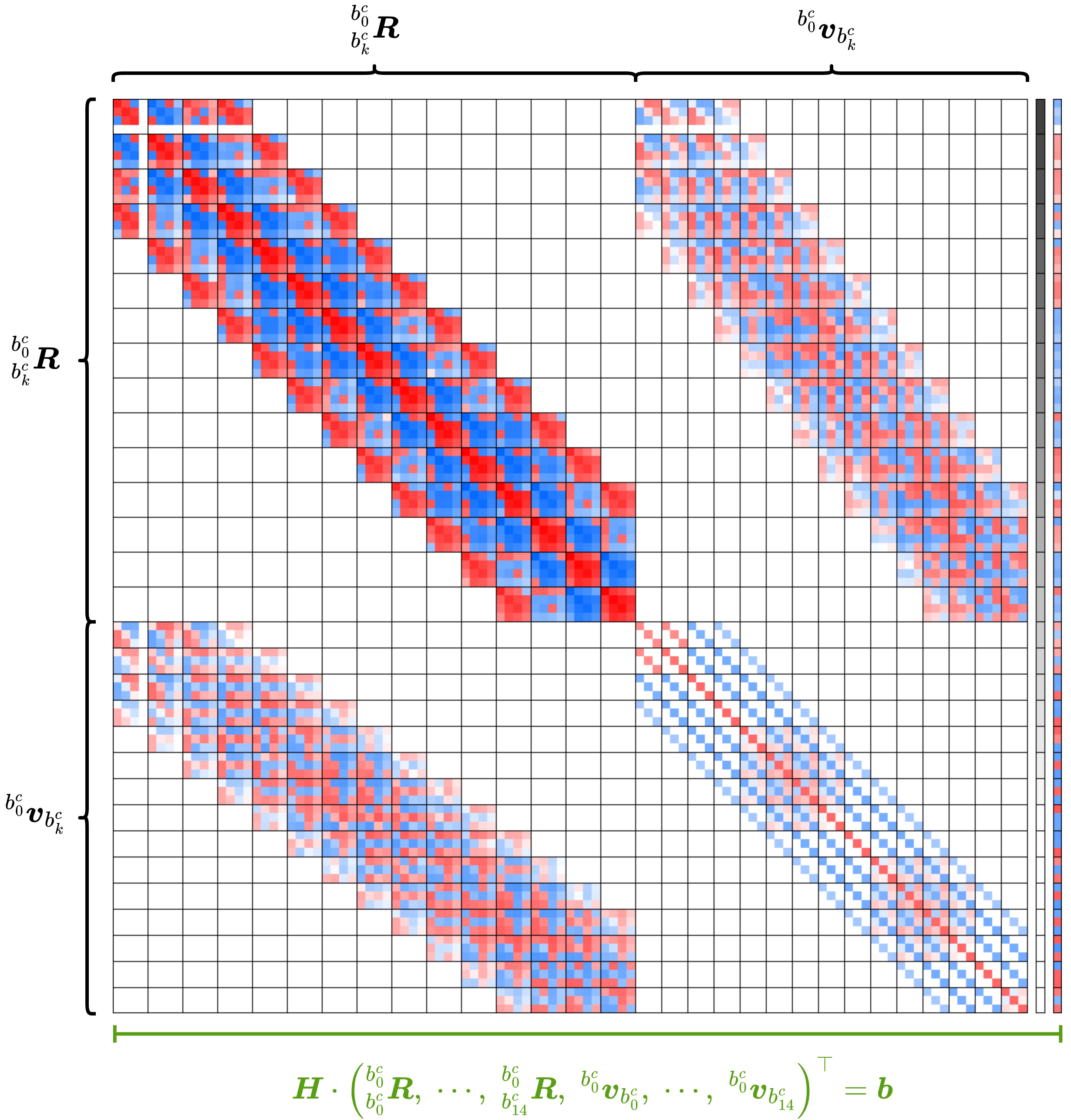}
			\caption{A visualization of the sparsity pattern of the normal equations built during one iteration of Levenberg-Marquardt.
				For better visibility, only 15 rotation and velocity control points are used to	generate this plot. Parameter blocks are separated by black lines, in which the rotation (represented as unit quaternion in the estimator) and velocity control points are four and three dimensional respectively. Darker reds and blues represent larger positive and negative values, while whites are zeros.}
			\label{fig:lm_equ}
		\end{figure}
	
		To evaluate the feasibility of the proposed method, we carried out the simulation tests, where three radars (denoted as \emph{RAD-1}, \emph{RAD-2}, and \emph{RAD-3}) and three IMUs (denoted as \emph{IMU-1}, \emph{IMU-2}, and \emph{IMU-3}) are simulated.
		All sensor measurements are generated with properly additive Gaussian noise.
		To avoid the parameter unobservability due to degenerate motions or ill-distributed radar targets, we constructed uniformly distributed targets and a sufficiently excited trajectory to simulate inertial and radar measurements, as shown in Fig. \ref{fig:simu_scene}.
		
		To better understand the processing of \emph{RIs-Calib} and evaluate its convergence performance, we plotted the root-mean-square errors (RMSEs) of spatiotemporal parameters and IMU biases in each iteration, as shown in Fig. \ref{fig:simu_opt_process}.
		As can be seen, the extrinsic rotations of IMUs are well estimated together with rotation B-spline recovery in \emph{INIT(1)}, followed by \emph{INIT(2)} where other extrinsics are initialized alongside the gravity.
		In \emph{INIT(3)}, these quantities are refined with the velocity B-spline recovery.
		At this point, all parameters in the estimator except the intrinsic and temporal parameters have been well initialized.
		Subsequently, three batch optimizations are performed successively, where a refinement strategy is employed to ensure the objective function reaches the global minimum efficiently.
		Specifically, we optimize spatial parameters, the gravity vector, and all control points in \emph{BO(1)}.
		In the following \emph{BO(2)} and \emph{BO(3)}, we sequentially add the temporal and intrinsic parameters into the estimator.
		Note that the final batch optimization, i.e., \emph{BO(3)}, is exactly a global optimization, where all parameters are included and optimized in the estimator to guarantee the global optimum.
		The final calibration accuracy can reach $1\;mm$ for translation, $0.05\;deg$ for rotation, and $0.1\;ms$ for time offset.
		As for IMU intrinsics, the accuracy of biases reaches $10^{-3}$ level for the accelerometer and $10^{-5}$ level for the gyroscope.
		These results demonstrate the excellent convergence performance and high calibration accuracy \emph{RIs-Calib} yields.
	
		To intuitively reflect the system sparsity benefited from the employed B-splines, we plot the normal equations in one Levenberg-Marquardt iteration, as shown in Fig. \ref{fig:lm_equ}.
		Although maintaining numerous control points in the estimator leads to a large system of equations, the symmetric information	matrix is sparsely populated due to the local controllability of B-splines.
		Therefore, sparse solvers, e.g., the \emph{sparse Schur solver} and \emph{sparse Cholesky solver}, could be employed to accelerate the computation when solving equations.
		Additionally, it is worth noting that the primary diagonal of the information matrix in Fig. \ref{fig:lm_equ} is populated by overlapping four-block-size matrices, which is related to the employed four-order uniform B-splines (i.e., three-degree ones) to represent the rotation and velocity curves.
		
		\section{Real-world Experiment}
		\begin{figure}[t]
			\centering
			\includegraphics[width=0.45\linewidth]{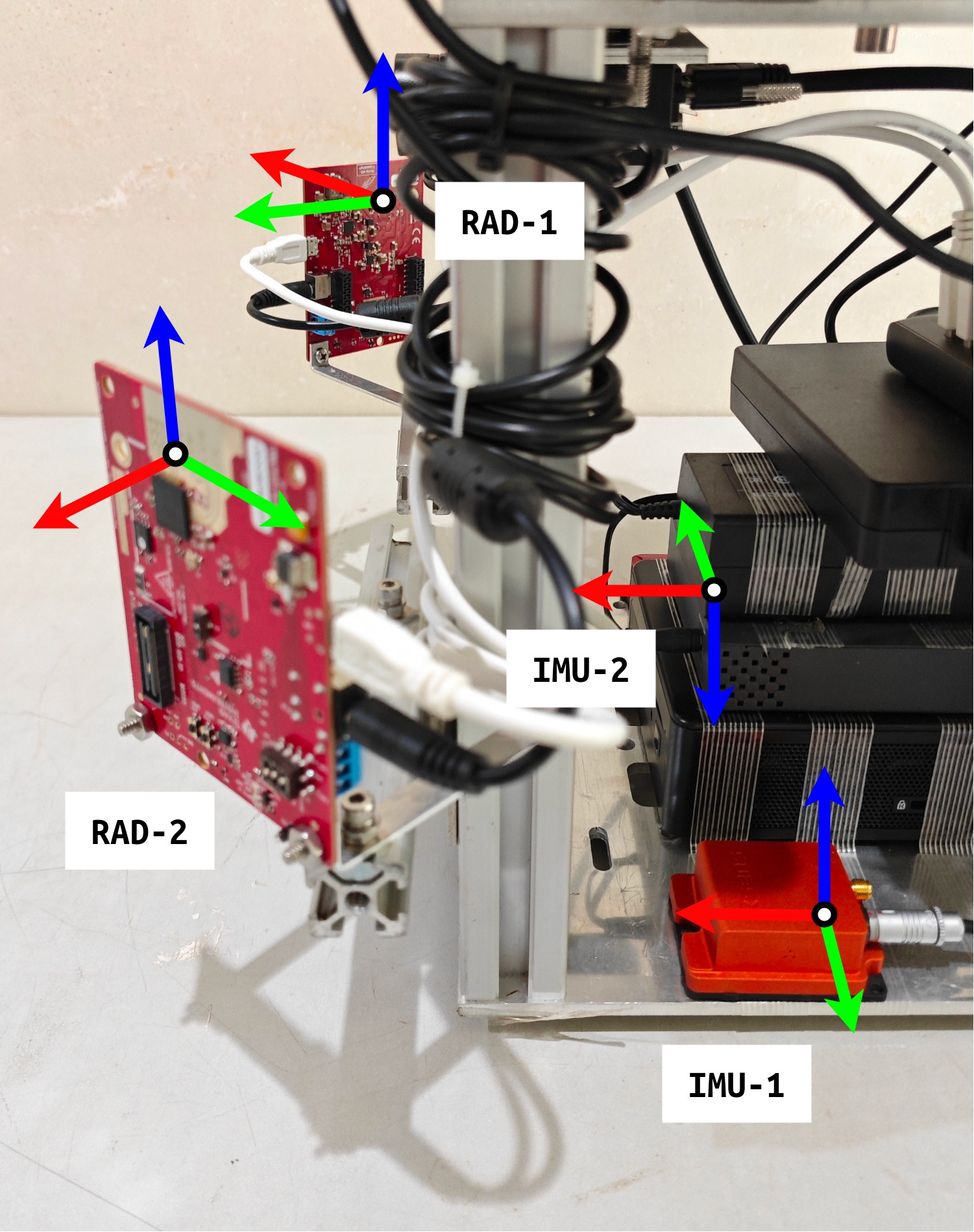}
			\includegraphics[width=0.45\linewidth]{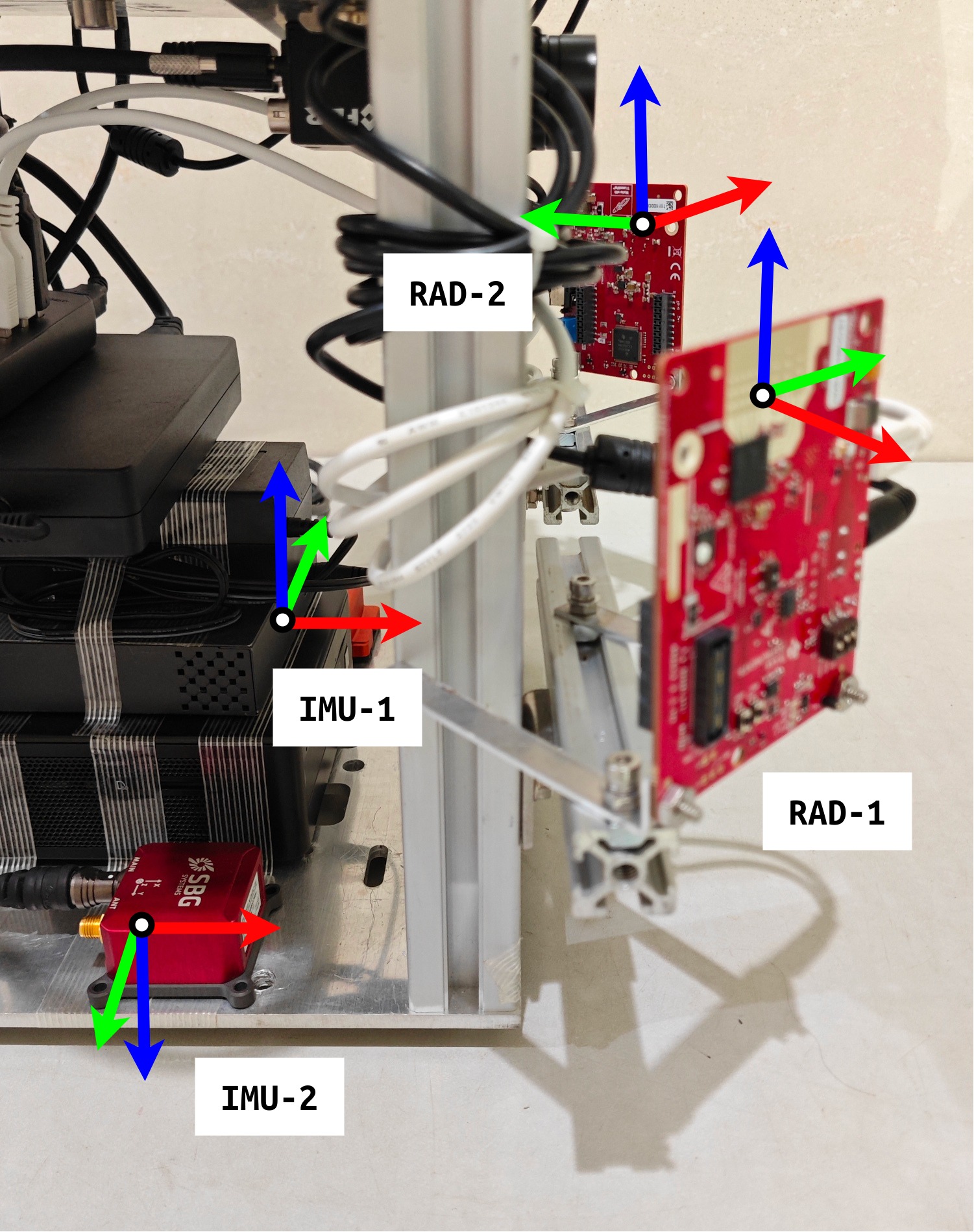}
			\label{fig:hardware}
				
			\vspace{1mm}

			\includegraphics[width=0.45\linewidth]{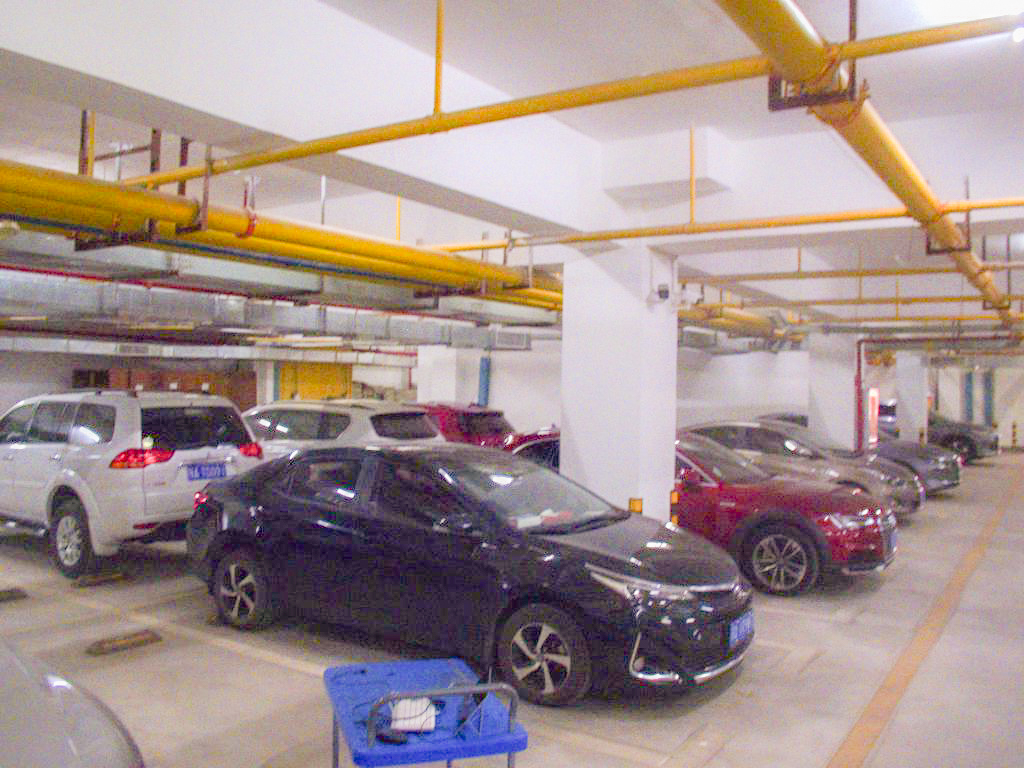}
			\includegraphics[width=0.45\linewidth]{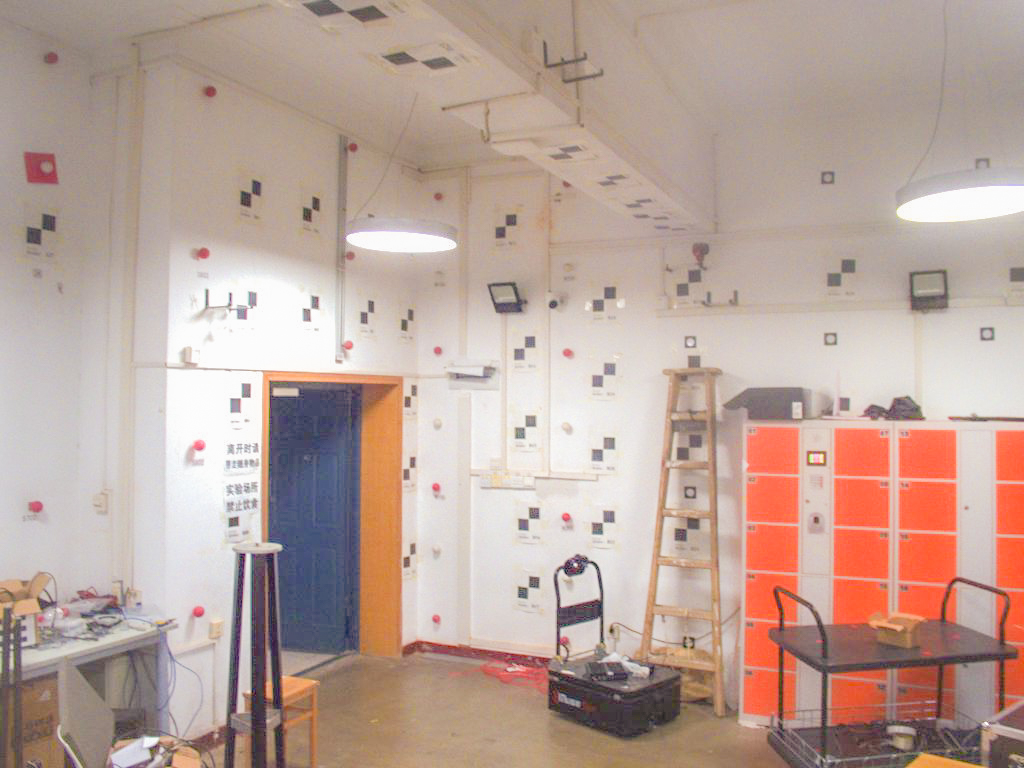}
			\label{fig:scenario}
			\caption{The hardware platform and the typical scenarios in real-world experiments.}
			
			\label{fig:real_world_exp}
		\end{figure}
		
	\begin{table*}[t!]
	\centering
	\begin{threeparttable}
	\caption{The spatiotemporal calibration results of \emph{RIs-Calib} and \emph{x-RIO} in Monte-Carlo real-world experiments}
	\label{tab:error_distribution}
	\begin{tabular}{c|c|c|rrrrrr|c}
	\toprule
	\multirow{4}{*}{Method}                                 & \multirow{4}{*}{Data}   & \multirow{4}{*}{Pair}    & \multicolumn{6}{c|}{Extrinsic}                                                                                                                                                                                                                                                                                             & Temporal                             \\ \cmidrule{4-10} 
	                                                        &                         &                          & \multicolumn{3}{c|}{Rotation Error $(deg)$}                                                                                                                         & \multicolumn{3}{c|}{Translation Error $(cm)$}                                                                                                        & Time Offset                          \\ \cmidrule{4-10} 
	                                                        &                         &                          & \multicolumn{1}{c}{$\delta{\boldsymbol{\theta}(r)}$} & \multicolumn{1}{c}{$\delta{\boldsymbol{\theta}(p)}$} & \multicolumn{1}{c|}{$\delta{\boldsymbol{\theta}(y)}$} & \multicolumn{1}{c}{$\delta{\boldsymbol{p}(x)}$} & \multicolumn{1}{c}{$\delta{\boldsymbol{p}(y)}$} & \multicolumn{1}{c|}{$\delta{\boldsymbol{p}(z)}$} & $\tau$ $(ms)$                        \\ \midrule\midrule
	\multirow{11}{*}{\rotatebox{90}{RIs-Calib (Ours)}}       & \multirow{3}{*}{Seq. 1} & $b^1\leftrightarrow b^2$ & \acrot{0.01}$\pm$0.01                                        & \acrot{0.02}$\pm$0.01                                        & \multicolumn{1}{r|}{\acrot{0.02}$\pm$0.02}                    & \actrans{0.03}$\pm$0.02                                   & \actrans{0.01}$\pm$0.01                                   & \actrans{0.05}$\pm$0.03                                    & \multicolumn{1}{r}{-38.27$\pm$0.44}  \\
	                                                        &                         & $b^1\leftrightarrow r^1$ & \acrot{0.74}$\pm$0.86                                        & \acrot{0.61}$\pm$0.95                                        & \multicolumn{1}{r|}{\acrot{0.51}$\pm$0.46}                    & \actrans{0.46}$\pm$0.31                                   & \actrans{0.38}$\pm$0.16                                   & \actrans{0.29}$\pm$0.17                                    & \multicolumn{1}{r}{-117.85$\pm$0.80} \\
	                                                        &                         & $b^1\leftrightarrow r^2$ & \acrot{0.44}$\pm$0.70                                        & \acrot{0.69}$\pm$0.97                                        & \multicolumn{1}{r|}{\acrot{0.41}$\pm$0.18}                    & \actrans{0.31}$\pm$0.20                                   & \actrans{0.46}$\pm$0.15                                   & \actrans{0.21}$\pm$0.19                                    & \multicolumn{1}{r}{-115.74$\pm$0.91} \\ \cmidrule{2-10} 
	                                                        & \multirow{3}{*}{Seq. 2} & $b^1\leftrightarrow b^2$ & \acrot{0.01}$\pm$0.01                                        & \acrot{0.01}$\pm$0.01                                        & \multicolumn{1}{r|}{\acrot{0.02}$\pm$0.01}                    & \actrans{0.01}$\pm$0.01                                   & \actrans{0.02}$\pm$0.01                                   & \actrans{0.03}$\pm$0.01                                    & \multicolumn{1}{r}{-37.19$\pm$0.16}  \\
	                                                        &                         & $b^1\leftrightarrow r^1$ & \acrot{0.40}$\pm$0.69                                        & \acrot{0.38}$\pm$0.24                                        & \multicolumn{1}{r|}{\acrot{0.30}$\pm$0.15}                    & \actrans{0.33}$\pm$0.08                                   & \actrans{0.21}$\pm$0.24                                   & \actrans{0.19}$\pm$0.23                                    & \multicolumn{1}{r}{-116.64$\pm$0.98} \\
	                                                        &                         & $b^1\leftrightarrow r^2$ & \acrot{0.17}$\pm$0.20                                        & \acrot{0.24}$\pm$0.16                                        & \multicolumn{1}{r|}{\acrot{0.18}$\pm$0.23}                    & \actrans{0.18}$\pm$0.19                                   & \actrans{0.26}$\pm$0.21                                   & \actrans{0.18}$\pm$0.07                                    & \multicolumn{1}{r}{-115.51$\pm$0.50} \\ \cmidrule{2-10} 
	                                                        & \multirow{3}{*}{Seq. 3} & $b^1\leftrightarrow b^2$ & \acrot{0.02}$\pm$0.01                                        & \acrot{0.01}$\pm$0.01                                        & \multicolumn{1}{r|}{\acrot{0.01}$\pm$0.02}                    & \actrans{0.01}$\pm$0.01                                   & \actrans{0.03}$\pm$0.02                                   & \actrans{0.02}$\pm$0.01                                    & \multicolumn{1}{r}{-41.91$\pm$0.33}  \\
	                                                        &                         & $b^1\leftrightarrow r^1$ & \acrot{0.76}$\pm$0.86                                        & \acrot{0.36}$\pm$0.34                                        & \multicolumn{1}{r|}{\acrot{0.42}$\pm$0.37}                    & \actrans{0.29}$\pm$0.10                                   & \actrans{0.17}$\pm$0.12                                   & \actrans{0.24}$\pm$0.24                                    & \multicolumn{1}{r}{-116.60$\pm$0.70} \\
	                                                        &                         & $b^1\leftrightarrow r^2$ & \acrot{0.37}$\pm$0.39                                        & \acrot{0.76}$\pm$0.77                                        & \multicolumn{1}{r|}{\acrot{0.21}$\pm$0.05}                    & \actrans{0.36}$\pm$0.11                                   & \actrans{0.34}$\pm$0.06                                   & \actrans{0.28}$\pm$0.29                                    & \multicolumn{1}{r}{-116.34$\pm$0.89} \\ \midrule
	\multirow{11}{*}{\rotatebox{90}{x-RIO \cite{doer2021x}}} & \multirow{3}{*}{Seq. 1} & $b^1\leftrightarrow b^2$ & \acrot{0.23}$\pm$0.31                                        & \acrot{0.35}$\pm$0.29                                        & \multicolumn{1}{r|}{\acrot{0.23}$\pm$0.18}                    & \actrans{0.91}$\pm$0.96                                   & \actrans{0.87}$\pm$0.82                                   & \actrans{0.73}$\pm$0.80                                    & \ding{55}                            \\
	                                                        &                         & $b^1\leftrightarrow r^1$ & \acrot{0.68}$\pm$0.94                                        & \acrot{0.67}$\pm$0.88                                        & \multicolumn{1}{r|}{\acrot{0.61}$\pm$0.79}                    & \actrans{1.23}$\pm$1.64                                   & \actrans{1.35}$\pm$0.72                                   & \actrans{1.40}$\pm$1.23                                    & \ding{55}                            \\
	                                                        &                         & $b^1\leftrightarrow r^2$ & \acrot{0.72}$\pm$0.74                                        & \acrot{0.61}$\pm$0.76                                        & \multicolumn{1}{r|}{\acrot{0.73}$\pm$0.86}                    & \actrans{0.98}$\pm$0.92                                   & \actrans{1.17}$\pm$1.39                                   & \actrans{1.01}$\pm$1.38                                    & \ding{55}                            \\ \cmidrule{2-10} 
	                                                        & \multirow{3}{*}{Seq. 2} & $b^1\leftrightarrow b^2$ & \acrot{0.30}$\pm$0.24                                        & \acrot{0.19}$\pm$0.18                                        & \multicolumn{1}{r|}{\acrot{0.27}$\pm$0.21}                    & \actrans{0.87}$\pm$0.72                                   & \actrans{1.08}$\pm$0.91                                   & \actrans{0.92}$\pm$0.98                                    & \ding{55}                            \\
	                                                        &                         & $b^1\leftrightarrow r^1$ & \acrot{0.75}$\pm$0.80                                        & \acrot{0.58}$\pm$0.92                                        & \multicolumn{1}{r|}{\acrot{0.63}$\pm$0.72}                    & \actrans{1.52}$\pm$1.47                                   & \actrans{1.20}$\pm$1.11                                   & \actrans{1.33}$\pm$1.07                                    & \ding{55}                            \\
	                                                        &                         & $b^1\leftrightarrow r^2$ & \acrot{0.62}$\pm$0.95                                        & \acrot{0.57}$\pm$0.72                                        & \multicolumn{1}{r|}{\acrot{0.66}$\pm$0.69}                    & \actrans{1.03}$\pm$0.94                                   & \actrans{1.12}$\pm$0.96                                   & \actrans{1.44}$\pm$1.26                                    & \ding{55}                            \\ \cmidrule{2-10} 
	                                                        & \multirow{3}{*}{Seq. 3} & $b^1\leftrightarrow b^2$ & \acrot{0.29}$\pm$0.27                                        & \acrot{0.32}$\pm$0.28                                        & \multicolumn{1}{r|}{\acrot{0.36}$\pm$0.30}                    & \actrans{0.71}$\pm$0.65                                   & \actrans{0.88}$\pm$0.83                                   & \actrans{0.90}$\pm$0.87                                    & \ding{55}                            \\
	                                                        &                         & $b^1\leftrightarrow r^1$ & \acrot{0.69}$\pm$0.54                                        & \acrot{0.56}$\pm$0.58                                        & \multicolumn{1}{r|}{\acrot{0.74}$\pm$0.75}                    & \actrans{1.34}$\pm$1.28                                   & \actrans{1.25}$\pm$1.09                                   & \actrans{1.14}$\pm$0.99                                    & \ding{55}                            \\
	                                                        &                         & $b^1\leftrightarrow r^2$ & \acrot{0.63}$\pm$0.70                                        & \acrot{0.78}$\pm$0.61                                        & \multicolumn{1}{r|}{\acrot{0.55}$\pm$0.89}                    & \actrans{1.21}$\pm$1.01                                   & \actrans{1.32}$\pm$1.17                                   & \actrans{1.25}$\pm$1.04                                    & \ding{55}                            \\ \bottomrule
	\end{tabular}
			\begin{tablenotes} 
			\item[*] Extrinsic translations in $(cm)$, extrinsic Euler angles in $(deg)$, and time offsets in $(ms)$.
			\item[*] Cells with darker colors indicate larger errors for the corresponding parameters.
			\end{tablenotes}
			\end{threeparttable}
	\end{table*}

		\begin{figure}[t!]
			\centering
			\includegraphics[width=0.95\linewidth]{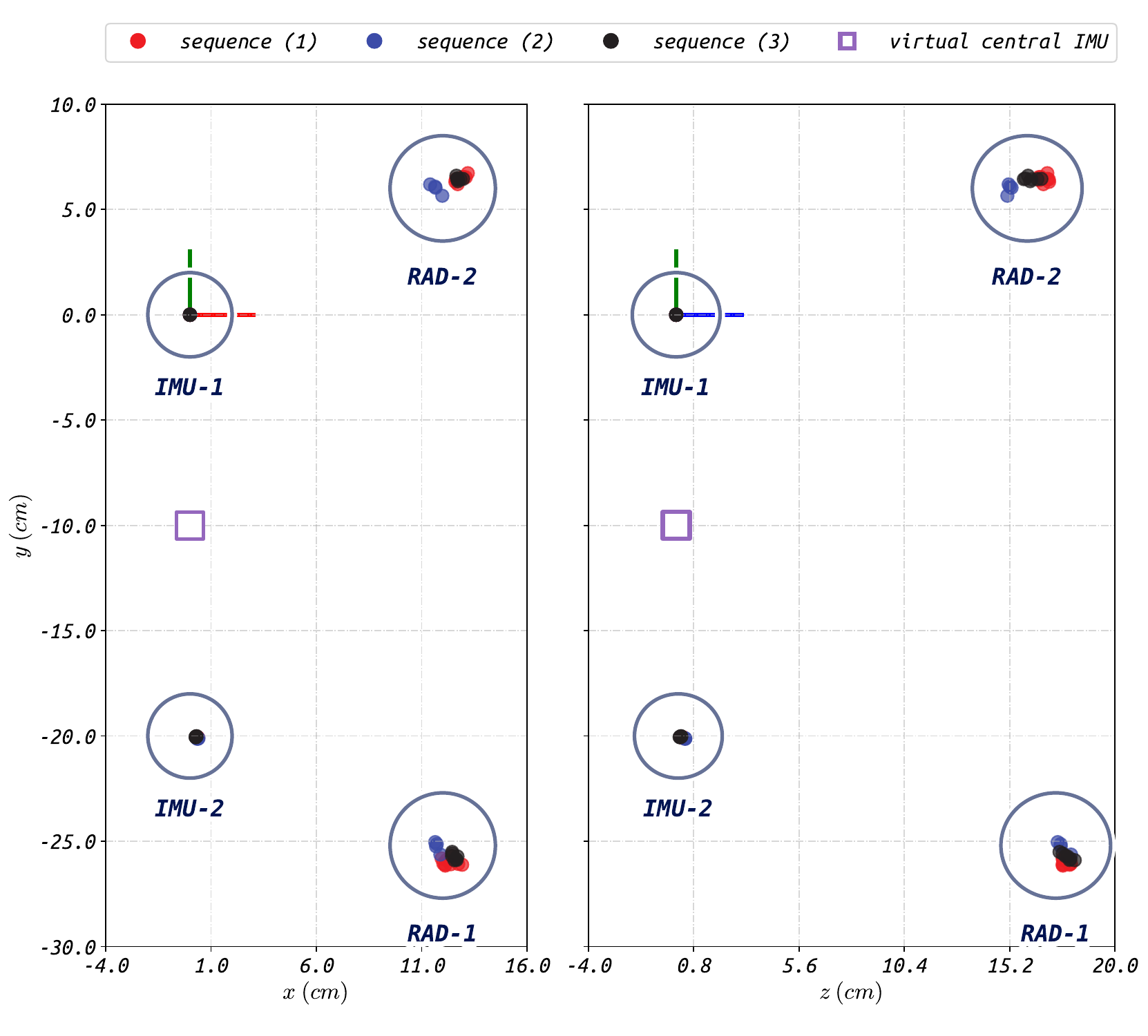}
			\caption{The distribution of extrinsic translations for two radars and two IMUs. All parameters are with respect to \emph{IMU-1}.}
			\label{fig:real_world_scatter}
		\end{figure}
		
		To comprehensively evaluate \emph{RIs-Calib}, we performed real-world experiments where the self-assembled hardware platform shown in Fig. \ref{fig:real_world_exp} is employed to collect data.
		Altogether two \emph{AWR1843BOOST} 3D radars (denoted as \emph{RAD-1} and \emph{RAD-2}), an \emph{XSens MTI-G-710} IMU (denoted as \emph{IMU-1}), and a \emph{SBG ELLIPSE-A} IMU (denoted as \emph{IMU-2}) are integrated into the sensor suite.
		The sampling rate of radars is set to 10 Hz, and for IMUs, it is 400 Hz (\emph{IMU-1}) and 200 Hz (\emph{IMU-2}).
		
		\subsection{Accuracy, Repeatability, and Convergence Evaluation}

		We first evaluated our proposed \emph{RIs-Calib} on calibration accuracy and repeatability.
		Altogether three data sequences collected in different scenarios were used, and each one of them was segmented into several data pieces lasting $50\;s$ for the Monte-Carlo tests.
		Fig. \ref{fig:real_world_scatter} depicts the calibration results of extrinsic translations of \emph{RIs-Calib}, all of which are expressed with respect to \emph{IMU-1}. The virtual central IMU is at the center of \emph{IMU-1} and \emph{IMU-2} due to the introduced center residuals.
		Intuitively, \emph{RIs-Calib} possesses good repeatability, especially for the extrinsic calibration of two IMUs, whose range is within $1\;mm$.
		This is due to the high-frequency and high-precision inertial measurements IMUs provide.
		In terms of radars, their repeatability is lower than IMUs' and the range is within $2\;cm$, which is reasonable since the accuracy of raw radar measurements is poor, especially the position observation of targets.
		
		The open-source radar-inertial odometry \emph{x-RIO} \cite{doer2021x} is then selected as a benchmark for calibration comparison.
		Since \emph{x-RIO} only supports online radar-IMU extrinsic calibration but not time offset calibration, we synchronized the radar data using the time offset estimated by \emph{RIs-Calib} prior to conducting the \emph{x-RIO} calibration to ensure a fair comparison.
		Additionally, we used the extrinsic calibration results from \emph{RIs-Calib} as the initial values for \emph{x-RIO}, as \emph{x-RIO}, being an odometry system with online calibration capability, requires well-initialized extrinsics to achieve convergence.
		Table \ref{tab:error_distribution} summarizes the calibration error distribution of spatiotemporal parameters for both \emph{RIs-Calib} and \emph{x-RIO}.
		Cells with darker colors indicate larger errors for the corresponding parameters (a deeper red signifies greater extrinsic rotational error, while a deeper purple indicates larger extrinsic translational error).
		The ground truth of extrinsics is obtained by computer-aided design (CAD). As for time offsets, considering the difficulty in obtaining the ground truth, calibration results are given by means and standard deviations (STDs) of estimates.
		As expected, the calibration results for extrinsic rotations and translations of IMUs are with higher accuracy and repeatability than those of radars, which holds for both \emph{RIs-Calib} and \emph{x-RIO}.
		In comparison, \emph{RIs-Calib} demonstrates overall superiority over \emph{x-RIO} in calibration accuracy, particularly evident in the calibration of extrinsic translations.
		The extrinsic errors of IMUs from \emph{RIs-Calib} are less than $0.02\;deg$ and $0.05\;cm$ with STDs within $0.02\;deg$ and $0.02\;cm$.
		As for radars, errors are $0.45\;deg$ and $0.30\;cm$ in average, and STDs are within $0.90\;deg$ and $0.50\;cm$.
		In terms of temporal parameters, since the synchronization of the platform is implemented on the software layer rather than the hardware layer, the time offsets could have subtle distinctions among separate startups, i.e., for three independently collected sequences.
		But in general, the STDs of time offsets calibrated by \emph{RIs-Calib} are less than $1\;ms$.
		These results demonstrate that \emph{RIs-Calib} could calibrate spatiotemporal parameters with high accuracy and excellent repeatability for multi-radar multi-IMU suites.
		
		
		\begin{figure}[t]
			\centering
			\includegraphics[width=\linewidth]{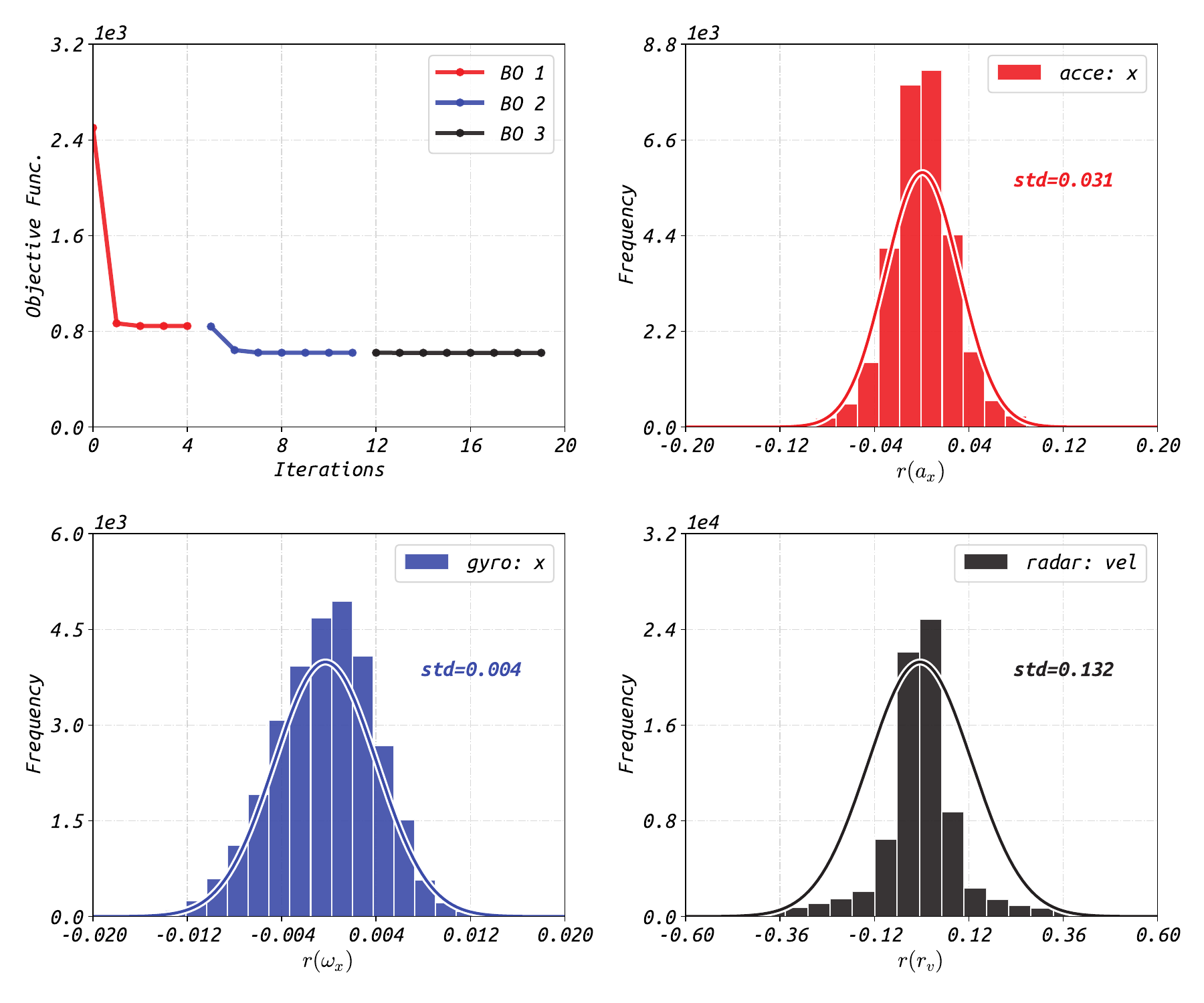}
			\caption{The variation of the objective function in three batch optimizations and the distributions of final residuals.}
			\label{fig:iter_info}
		\end{figure}
			
		\begin{table}[t!]
			\centering
			\begin{threeparttable}
			\caption{The multi-IMU spatiotemporal calibration results from \emph{RIs-Calib} and \emph{Kalibr}}
			\label{tab:multi_imu_calib}
		\begin{tabular}{cccc}
				\midrule\midrule
				Rot. Error                        & $\delta{\boldsymbol{\theta}(r)}$ $(deg)$ & $\delta{\boldsymbol{\theta}(p)}$ $(deg)$ & $\delta{\boldsymbol{\theta}(y)}$ $(deg)$ \\ \midrule
				\multicolumn{1}{c|}{RAD-1-Aided}  & \acrot{0.03}$\pm$0.02                            & \acrot{0.04}$\pm$0.02                            & \acrot{0.05}$\pm$0.02                            \\
				\multicolumn{1}{c|}{RAD-2-Aided}  & \acrot{0.04}$\pm$0.03                            & \acrot{0.02}$\pm$0.02                            & \acrot{0.03}$\pm$0.03                            \\
				\multicolumn{1}{c|}{Radars-Aided} & \acrot{0.02}$\pm$0.01                            & \acrot{0.02}$\pm$0.02                            & \acrot{0.03}$\pm$0.02                            \\
				\multicolumn{1}{c|}{kalibr}       & \acrot{0.01}$\pm$0.01                            & \acrot{0.01}$\pm$0.01                            & \acrot{0.01}$\pm$0.01                            \\ \midrule\midrule
				Trans. Error                      & $\delta{\boldsymbol{p}(x)}$ $(mm)$       & $\delta{\boldsymbol{p}(y)}$ $(mm)$       & $\delta{\boldsymbol{p}(z)}$ $(mm)$       \\ \midrule
				\multicolumn{1}{c|}{RAD-1-Aided}  & \actrans{0.41}$\pm$0.52                            & \actrans{0.67}$\pm$0.53                            & \actrans{0.54}$\pm$0.65                            \\
				\multicolumn{1}{c|}{RAD-2-Aided}  & \actrans{0.52}$\pm$0.68                            & \actrans{0.59}$\pm$0.58                            & \actrans{0.44}$\pm$0.51                            \\
				\multicolumn{1}{c|}{Radars-Aided} & \actrans{0.38}$\pm$0.51                            & \actrans{0.20}$\pm$0.54                            & \actrans{0.38}$\pm$0.57                            \\
				\multicolumn{1}{c|}{kalibr}       & \actrans{0.24}$\pm$0.28                            & \actrans{0.12}$\pm$0.10                            & \actrans{0.13}$\pm$0.06                            \\ \midrule\midrule
				Time Est.                         & \multicolumn{3}{c}{$\tau$ $(ms)$}                                                                                              \\ \midrule
				RAD-1-Aided                       & RAD-2-Aided                              & Radars-Aided                             & kalibr                                   \\ \midrule
				38.06$\pm$0.62                    & 37.52$\pm$0.79                           & 37.96$\pm$0.43                           & 37.88$\pm$0.11                           \\ \midrule\midrule
				\end{tabular}
		\begin{tablenotes} 
		\item[*] Extrinsic translations in $(mm)$, extrinsic Euler angles in $(deg)$, and time offsets in $(ms)$.
		\item[*] Cells with darker colors indicate larger errors for the corresponding parameters.
		\end{tablenotes}
		\end{threeparttable}
		\end{table}
		
		Fig. \ref{fig:iter_info} shows ($i$) the convergence performance of \emph{RIs-Calib} in batch optimizations, and ($ii$) the residual distributions after optimization.
		In the first batch optimization, the objective function is reduced significantly, and the coarse B-splines recovered in initialization are refined to better states alongside extrinsic optimization.
		In the next optimization, the time offsets were optimized, which resulted in the objective function reducing notably again.
		In the final optimization, the intrinsics of IMUs were involved and optimized with other parameters to ensure the objective function reaches the global minimum.
		In general, \emph{RIs-Calib} could converge within $20$ iterations.
		The residuals after three optimizations obey the zero-mean Gaussian distribution, which clearly indicates that \emph{RIs-Calib} is able to estimate spatiotemporal parameters effectively without bias.
				
		\subsection{Multi-IMU Calibration Comparison with Kalibr}
		
		Quantitative comparisons for spatiotemporal calibration between \emph{RIs-Calib} and other well-established baselines are subsequently conducted.
		Considering the lack of open-sourcing radar-IMU spatiotemporal calibration works, we regard \emph{RIs-Calib} as a radar-aided muti-IMU calibration method and compare it with camera-aided muti-IMU calibration in the well-known \emph{Kalibr} \cite{rehder2016extending}.
		The corresponding results are summarized in Table \ref{tab:multi_imu_calib}, where both single-radar-aided and multi-radar-aided multi-IMU calibrations are evaluated. 
		It can be found that the camera-aided \emph{Kalibr} achieves the highest calibration accuracy.
		The errors of extrinsic rotation and translation from \emph{Kalibr} are less than $0.01\;deg$ and $0.25\;mm$ respectively, which are smaller than those from \emph{RIs-Calib}.
		This is mainly due to the more noisy target measurements from radars compared with the images from the camera.
		Nonetheless, \emph{RIs-Calib} still can provide acceptable spatiotemporal parameters.
		Meanwhile, as a targetless calibration method, \emph{RIs-Calib} has stronger flexibility and usability compared to the target-based (chessboard-based) \emph{Kalibr}.
		Furthermore, it can be seen that multi-radar-aided multi-IMU calibration can provide more accurate spatiotemporal parameters than single-radar-aided one, which indicates that involving multiple radars could effectively suppress the noise of radar in radar-aided multi-IMU spatiotemporal calibration.
		
		\begin{figure}[t!]
			\centering
			\includegraphics[width=\linewidth]{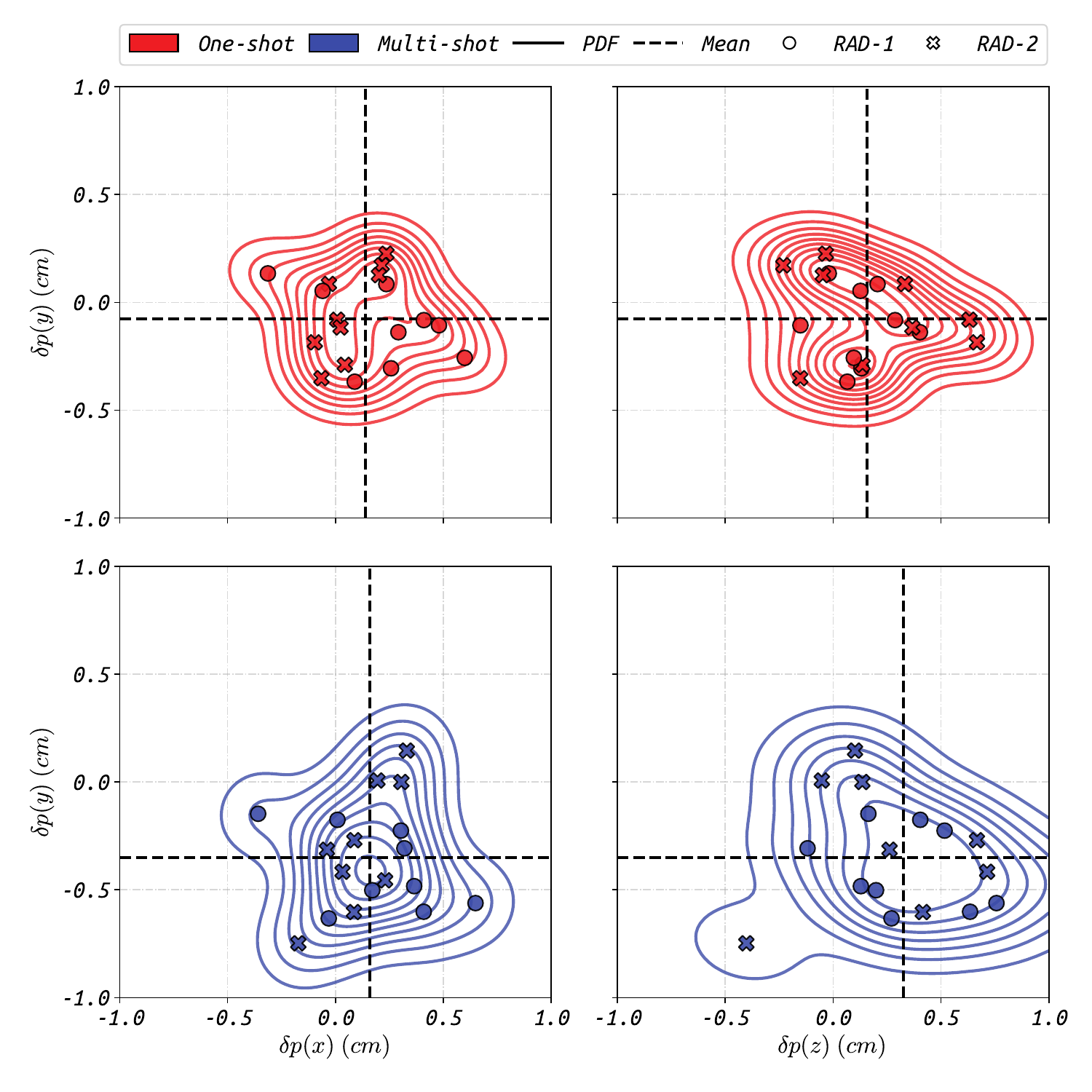}
			\caption{The error distribution for extrinsic translations of two radars in different calibration settings: ($i$) One-shot: spatiotemporal parameters of \emph{RAD-1}, \emph{RAD-2}, and \emph{IMU-1} are jointly optimized in a multi-radar multi-IMU calibration, ($ii$) Multi-shot: multiple single-radar single-IMU spatiotemporal calibrations where two radars are calibrated separately. All parameters are with respect to \emph{IMU-1}.}
			\label{fig:one_multi_shot_trans}
		\end{figure}

		\begin{table*}[t!]
		\centering
			\caption{The computational efficiency statistics of \emph{RIs-Calib} in Monte-Carlo real-world experiments}
			\label{tab:config_tests}
		\begin{tabular}{c|ccccccccccc}
		\midrule\midrule
		\multirow{2}{*}[-1ex]{Config.} & \multicolumn{1}{l}{OS Name}                        & \multicolumn{2}{l|}{Ubuntu 20.04.6 LTS} & Processor                  & \multicolumn{7}{l}{12th Gen Intel® Core™ i9-12900H × 20}                                               \\ \cmidrule{2-12} 
& \multicolumn{1}{l}{OS Type}                        & \multicolumn{2}{l|}{64-bit}             & Graphics                   & \multicolumn{7}{c}{Mesa Intel® Graphics (ADL GT2) / Mesa Intel® Graphics (ADL GT2)}                    \\ \midrule\midrule
		\multirow{2}{*}{Dataset} & \multicolumn{1}{c|}{\multirow{2}{*}{Duration (s)}} & \multicolumn{3}{c|}{Factor Count}                                    & \multicolumn{7}{c}{Time Elapsed (s)}                                                                   \\ \cmidrule{3-12} 
& \multicolumn{1}{c|}{}                              & Acce.              & Gyro.              & \multicolumn{1}{c|}{Radar} & INIT(1) & INIT(2) & \multicolumn{1}{c|}{INIT(3)} & BO(1) & BO(2) & \multicolumn{1}{c|}{BO(3)} & Total  \\ \midrule
		Seq. 1                   & \multicolumn{1}{c|}{50}                            & 30k                & 30k                & \multicolumn{1}{c|}{70k}   & 0.461   & 0.233   & \multicolumn{1}{c|}{1.090}   & 1.861 & 5.687 & \multicolumn{1}{c|}{6.536} & 15.868 \\
		Seq. 2                   & \multicolumn{1}{c|}{50}                            & 30k                & 30k                & \multicolumn{1}{c|}{76k}   & 0.445   & 0.221   & \multicolumn{1}{c|}{1.169}   & 2.111 & 7.105 & \multicolumn{1}{c|}{6.763} & 17.814 \\
		Seq. 3                   & \multicolumn{1}{c|}{50}                            & 30k                & 30k                & \multicolumn{1}{c|}{86k}   & 0.475   & 0.222   & \multicolumn{1}{c|}{1.288}   & 2.344 & 7.004 & \multicolumn{1}{c|}{7.123} & 18.456 \\ \midrule\midrule
		\end{tabular}
		\end{table*}

		\begin{figure}[t!]
			\centering
			\includegraphics[width=\linewidth]{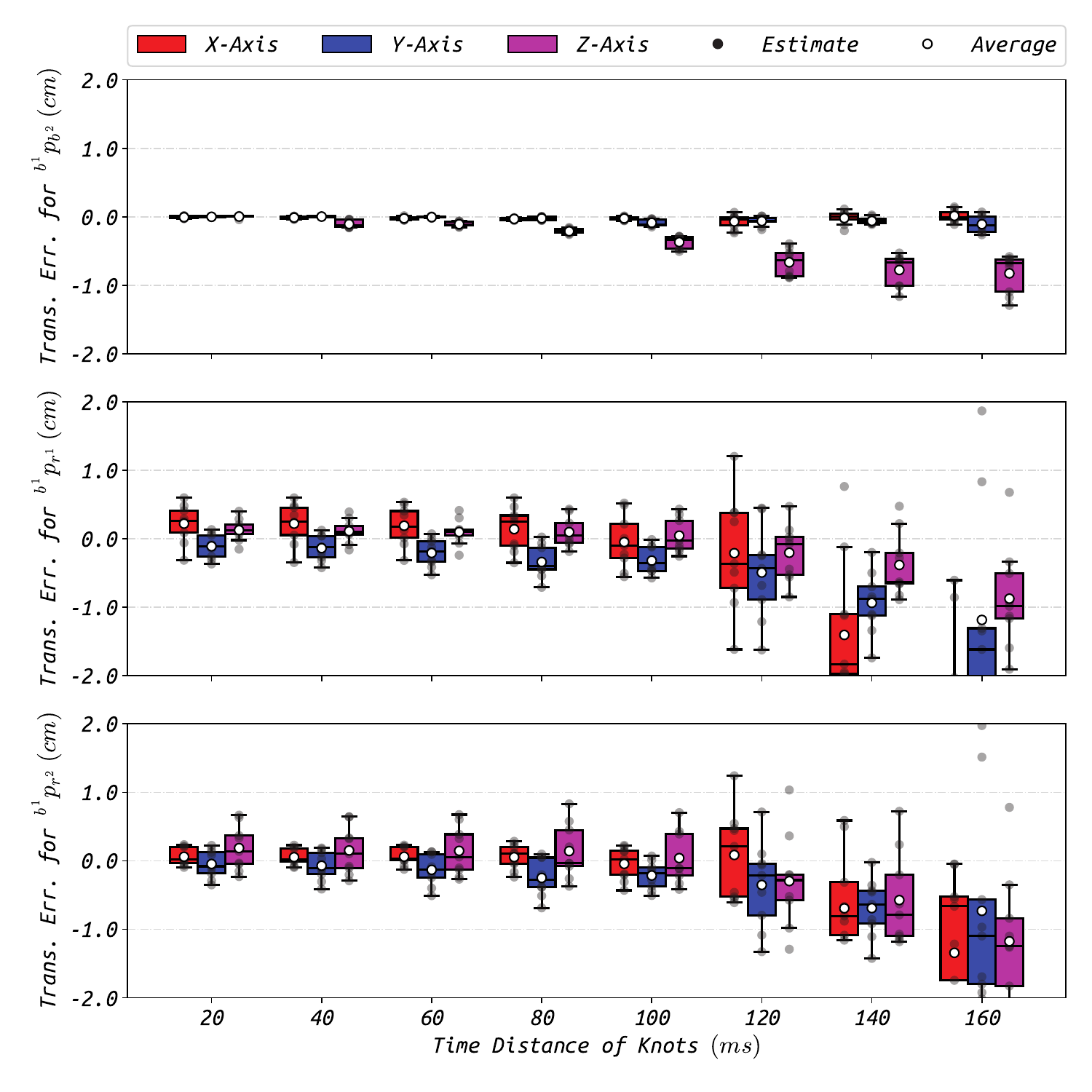}
			\caption{The error distribution for extrinsic translations in different settings of knot time distance varying from 20 ms to 180 ms. All parameters are with respect to \emph{IMU-1}.}
			\label{fig:time_dist}
		\end{figure}
		
		\subsection{Comparison of Joint and Separate Calibration}
		
		The proposed \emph{RIs-Calib} supports spatiotemporal calibration for radar-inertial suites that integrate any number of sensors.
		To explore the impact of different sensor configurations on the final calibration accuracy, one-shot (multi-radar multi-IMU) and multi-shot (multiple separate single-radar single-IMU) spatiotemporal calibrations are performed, where solving settings remain the same.
		Fig. \ref{fig:one_multi_shot_trans} shows the errors of extrinsic translations for two radars.
		The probability density function (PDFs) of translation errors from one-shot calibration are distributed more gathered and closer to zeros than the multi-shot one, which holds for both radars, indicating that joint optimization achieves more accurate and precise results.
		This is mainly due to that fusing noisy but sufficient measurements from two radars could recover accurate rotation and velocity B-splines and thus benefit the final spatiotemporal calibration.
		Meanwhile, compared with multiple separate calibrations, one-shot joint calibration is less labor-intensive and could guarantee more consistent results.
		
		\subsection{Evaluation of B-spline Representation}
		
		In most continuous-time-based calibration, the accuracy of spatiotemporal parameters is largely relevant to the adequacy of continuous-time representation.
		Take the uniform B-spline representation employed in this work as an example, smaller time distance of knots leads to more expressive B-splines. 
		Fig. \ref{fig:time_dist} shows the calibration results of extrinsic translations in Monte-Carlo experiments with different time distance settings for B-spline knots.
		It can be found that employing B-splines with a time distance less than $100\;ms$ could achieve acceptable spatiotemporal calibration accuracy.
		When utilizing B-splines whose time distance is larger than $100\;ms$, both accuracy and repeatability of spatiotemporal calibration would significantly reduce.
		Note that although a smaller time distance results in higher-performance calibration, more computational consumption is required, which should be carefully considered in practice based on the measurement frequency of sensors and intensity of motion excitation.

		\begin{figure}[t]
			\centering
			\includegraphics[width=\linewidth]{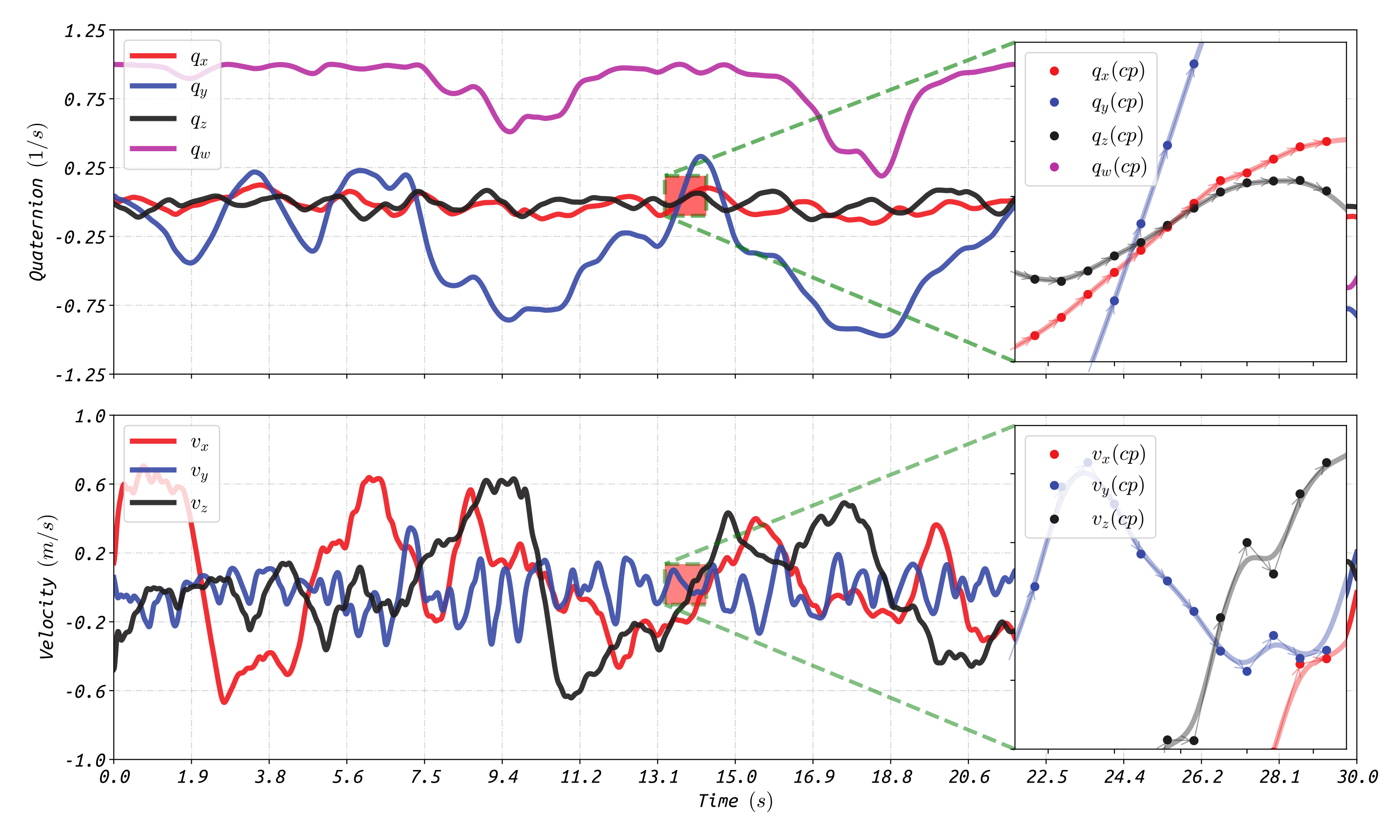}
			\caption{The estimated rotation and velocity B-splines, and their corresponding control points.}
			\label{fig:splines}
		\end{figure}
		
		The rotation and velocity B-splines estimated by \emph{RIs-Calib} in one Monte-Carlo test are shown in Fig. \ref{fig:splines}. The duration of the B-splines is intercepted to $30\;s$ for better readability, and the time distance of knots is set as $0.08\;s$.
		As can be seen, the control points are not exactly on curves, since the B-splines do not interpolate but only approximate the control points.
		Meanwhile, compared with the rotation B-splines, the velocity B-splines have more drastic variation trends.
		This is reasonable since the linear velocity is a higher-order quantity compared to the rigid motion.
		
		\subsection{Statistics of Computational Efficiency}
		
		To analyze the computational efficiency of \emph{RIs-Calib}, we statistics the average elapsed time in initialization and batch optimizations for three sequences, as shown in Table \ref{tab:config_tests}.
		Note that each solving is performed on ten threads and employs B-splines with a time distance of $0.08\;s$ between knots for both rotation and velocity representation.
		It could be found that the solving can be finished within $20\;s$, in which about $10\;\%$ cost by initialization procedure and $90\;\%$ by three batch optimizations.
		Additionally, as the number of radar measurements increases, the corresponding computation time grows reasonably.
	
		\section{Conclusion}
		In this work, we propose a targetless spatiotemporal calibrator termed as \emph{RIs-Calib} for multiple 3D radars and IMUs based on continuous-time batch estimation, which ($i$) supports spatial, temporal, and intrinsic calibration, ($ii$) requires no additional artificial infrastructure or prior knowledge.
		We perform a rigorous initialization procedure first to obtain initial guesses of states, followed by several batch optimizations to guarantee the global optimum of states.
		We carried out both simulated and real-world experiments to evaluate \emph{RIs-Calib}, and the results demonstrate its high accuracy and repeatability.
		Future work would focus on the improvements in the efficiency of \emph{RIs-Calib} and make it a real-time application.
		
		\section*{ACKNOWLEDGMENT}
		The calibration data acquisition is performed on the GREAT
		(GNSS+ REsearch, Application and Teaching) software, developed by the GREAT Group, School of Geodesy and Geomatics (SGG), Wuhan University.
		
		\section*{CRediT Authorship Contribution Statement}
		\textbf{Shuolong Chen}: Conceptualisation, Methodology, Software, Original Draft.
		\textbf{Xingxing Li}: Resources, Supervision, Review and Editing, Funding acquisition.
		\textbf{Shengyu Li}: Investigation, Formal analysis, Review and Editing.
		\textbf{Yuxuan Zhou} and \textbf{Shiwen Wang}: Data curation, Review and Editing.
	
	\bibliographystyle{IEEEtran}
	\bibliography{reference}
	\newpage
	\begin{IEEEbiography}[{\includegraphics[width=1in,height=1.25in,clip,keepaspectratio,cframe={black!8!white}]{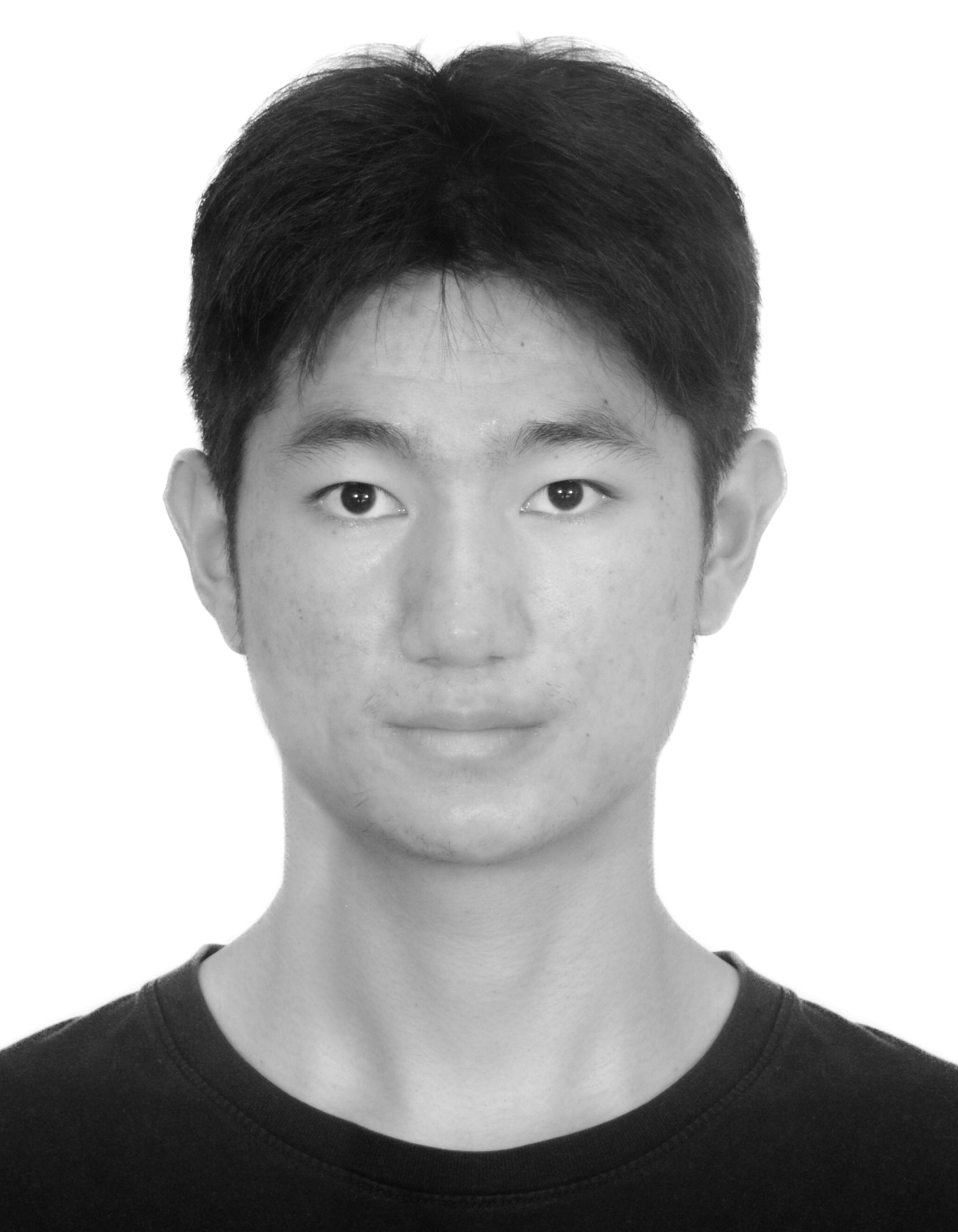}}]{Shuolong Chen}
		received the B.S. degree in geodesy and geomatics engineering from Wuhan University, Wuhan China, in 2023.
			
			He is currently a master candidate at the School of Geodesy and Geomatics (SGG), Wuhan University. His area of research currently focuses on integrated navigation systems and multi-sensor fusion, mainly on spatiotemporal calibration.
			
			Contact him via	e-mail: \emph{shlchen@whu.edu.cn}
	\end{IEEEbiography}
	\vspace{-8.0cm}
	\begin{IEEEbiography}[{\includegraphics[width=1in,height=1.25in,clip,keepaspectratio,cframe={black!8!white}]{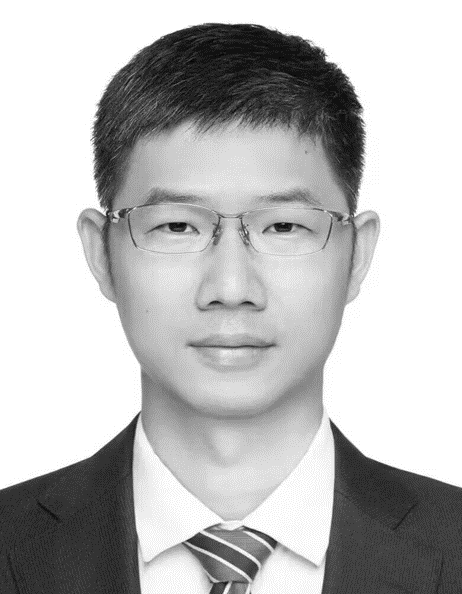}}]{Xingxing Li}
		received the Ph.D. degree in geodesy from Technical University of Berlin, Berlin, Germany, in 2015.
		
		He is currently a professor at the Wuhan University. His current research mainly involves GNSS precise data processing and multi-sensor fusion.
	\end{IEEEbiography}
	\vspace{-8.0cm}	
	\begin{IEEEbiography}[{\includegraphics[width=1in,height=1.25in,clip,keepaspectratio]{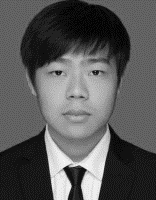}}]{Shengyu Li}
		received the M.S. degree in	geodesy and survey engineering from Wuhan University, Wuhan, China, in 2022.
		
		He is currently a doctor candidate at the school of Geodesy and Geomatics, Wuhan University, China. His current research focuses on multi-sensor fusion and integrated navigation system.
	\end{IEEEbiography}
	\vspace{-8.0cm}
	\begin{IEEEbiography}[{\includegraphics[width=1in,height=1.25in,clip,keepaspectratio]{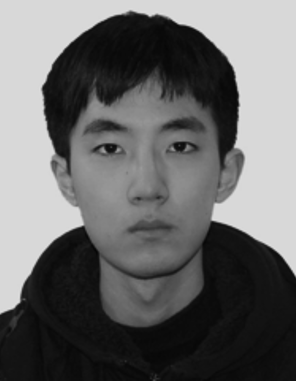}}]{Yuxuan Zhou}
		received the M.S. degree in geodesy and survey engineering from Wuhan University, Wuhan, China, in 2022.
		
		He is currently a doctor candidate at the school of Geodesy and Geomatics, Wuhan University, China. His current research focuses on
		vision-based mapping.
	\end{IEEEbiography}
	\vspace{-8.0cm}
	\begin{IEEEbiography}[{\includegraphics[width=1in,height=1.25in,clip,keepaspectratio]{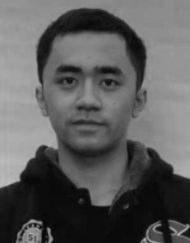}}]{Shiwen Wang}
		received the B.S. degree in navigation engineering from Wuhan University, Wuhan China, in 2021.
		
		He is currently a master candidate at the school of Geodesy and Geomatics, Wuhan University. His current research focuses on integrated navigation system and GNSS precise
		positioning.
	\end{IEEEbiography}

\end{document}